\documentclass[10pt]{amsart}

\usepackage[foot]{amsaddr}

\usepackage{amsthm}

\usepackage{xstring} 
\usepackage{floatrow, titletoc}
\usepackage{wrapfig}
\usepackage{booktabs} %

\usepackage[utf8]{inputenc} 
\usepackage{amsthm,amsmath, amssymb,bm, amsfonts, cases, mathtools, thmtools}
\usepackage{verbatim}
\usepackage{graphicx}
\graphicspath{{figures/}}
\usepackage{multicol}
\usepackage[usenames,dvipsnames]{xcolor}
\usepackage{algorithm}
\usepackage{algorithmicx}
\usepackage[noend]{algpseudocode}

\usepackage[%
    minnames=1,maxnames=99,maxcitenames=3,
    style=authoryear-comp, %
    citestyle=authoryear,
    sortcites=true,
    doi=false,url=false,
    isbn=false,
    dashed=false,
    hyperref,natbib,backend=biber]{biblatex}
    
\renewbibmacro{in:}{%
  \ifentrytype{article}{}{\printtext{\bibstring{in}\addspace}}}%
\bibliography{biblio}

\DeclareNameAlias{sortname}{first-last}

\DeclareCiteCommand{\cite}
  {\usebibmacro{prenote}}
  {\usebibmacro{citeindex}%
   \printtext[bibhyperref]{\usebibmacro{cite}}}
  {\multicitedelim}
  {\usebibmacro{postnote}}

\DeclareCiteCommand*{\cite}
  {\usebibmacro{prenote}}
  {\usebibmacro{citeindex}%
   \printtext[bibhyperref]{\usebibmacro{citeyear}}}
  {\multicitedelim}
  {\usebibmacro{postnote}}

\DeclareCiteCommand{\parencite}[\mkbibparens]
  {\usebibmacro{prenote}}
  {\usebibmacro{citeindex}%
    \printtext[bibhyperref]{\usebibmacro{cite}}}
  {\multicitedelim}
  {\usebibmacro{postnote}}

\DeclareCiteCommand*{\parencite}[\mkbibparens]
  {\usebibmacro{prenote}}
  {\usebibmacro{citeindex}%
    \printtext[bibhyperref]{\usebibmacro{citeyear}}}
  {\multicitedelim}
  {\usebibmacro{postnote}}

\DeclareCiteCommand{\footcite}[\mkbibfootnote]
  {\usebibmacro{prenote}}
  {\usebibmacro{citeindex}%
  \printtext[bibhyperref]{ \usebibmacro{cite}}}
  {\multicitedelim}
  {\usebibmacro{postnote}}

\DeclareCiteCommand{\footcitetext}[\mkbibfootnotetext]
  {\usebibmacro{prenote}}
  {\usebibmacro{citeindex}%
   \printtext[bibhyperref]{\usebibmacro{cite}}}
  {\multicitedelim}
  {\usebibmacro{postnote}}

\DeclareCiteCommand{\textcite}
  {\boolfalse{cbx:parens}}
  {\usebibmacro{citeindex}%
   \printtext[bibhyperref]{\usebibmacro{textcite}}}
  {\ifbool{cbx:parens}
     {\bibcloseparen\global\boolfalse{cbx:parens}}
     {}%
   \multicitedelim}
  {\usebibmacro{textcite:postnote}}

\usepackage{datetime}

\DeclareMathAlphabet\EuRoman{U}{eur}{m}{n}
\SetMathAlphabet\EuRoman{bold}{U}{eur}{b}{n}

\makeatletter
\let\reftagform@=\tagform@
\def\tagform@#1{\maketag@@@{\ignorespaces\textcolor{gray}{(#1)}\unskip\@@italiccorr}}
\renewcommand{\eqref}[1]{\textup{\reftagform@{\ref{#1}}}}
\makeatother

\declaretheorem[style=plain,numberwithin=section,name=Theorem]{theorem}
\declaretheorem[style=plain,sibling=theorem,name=Lemma]{lemma}

\declaretheorem[style=plain,sibling=theorem,name=Proposition]{proposition}

\declaretheorem[style=definition,qed=$\triangleleft$,sibling=theorem,name=Example]{example}
\numberwithin{theorem}{section}

\def\[#1\]{\begin{align}#1\end{align}}
\def\*[#1\]{\begin{align*}#1\end{align*}}

\def\clap#1{\hbox to 0pt{\hss#1\hss}}

\newcommand{\defas}{\vcentcolon=}  %

\newcommand{\datatrain}{S}

\newcommand{\dist}{\ \sim\ }

\newcommand{\Reals}{\mathbb{R}}

\newcommand{\Nats}{\mathbb{N}}

\newcommand{\NNReals}{\Reals_+}

\newcommand{\grad}{\nabla}

\newcommand{\dee}{\mathrm{d}}

\DeclareMathOperator{\sign}{sign}

\DeclareMathOperator*{\newlim}{\mathrm{lim}\vphantom{\mathrm{infsup}}}
\DeclareMathOperator*{\newmin}{\mathrm{min}\vphantom{\mathrm{infsup}}}
\DeclareMathOperator*{\newmax}{\mathrm{max}\vphantom{\mathrm{infsup}}}
\DeclareMathOperator*{\newinf}{\mathrm{inf}\vphantom{\mathrm{infsup}}}

\renewcommand{\lim}{\newlim}
\renewcommand{\min}{\newmin}
\renewcommand{\max}{\newmax}
\renewcommand{\inf}{\newinf}

\newcommand{\Measures}[1]{\mathcal{M}(#1)}
\newcommand{\ProbMeasures}[1]{\mathcal{M}_1(#1)}

\renewcommand{\Pr}{\mathbb{P}}
\def\EE{\mathbb{E}}

\newcommand{\defn}[1]{\emph{#1}}

\newcommand{\floor}[1]{\lfloor #1 \rfloor}

\newcommand{\cF}{\mathcal F}

\newcommand{\KLname}{\mathrm{KL}}

\newcommand{\KL}[3][]{\KLname#1(#2 #1|#1|#3 #1)}
\newcommand{\KLnamebin}{\mathrm{kl}}
\newcommand{\KLbin}[3][]{\KLnamebin#1(#2 #1|#1|#3 #1)}

\newcommand{\Normal}[2]{\mathcal N(#1,#2)}
\newcommand{\trace}{\mathrm{tr}}
\newcommand{\Bernoulli}[1]{\mathcal B_{#1}}

\newcommand{\loss}{\ell}

\newcommand{\pdim}{p}

\renewcommand{\algorithmicrequire}{\textbf{Input:}}
\renewcommand{\algorithmicensure}{\textbf{Output:}}
\newcommand{\e}{\mathrm{e}}

\newcommand{\Dist}{\mathcal D}

\newcommand{\EEE}[1]{\underset{#1}{\EE}}
\newcommand{\PPr}[1]{\underset{#1}{\Pr}}

\newcommand{\set}[2][]{#1\{#2 #1\}}

\newcommand{\abs}[2][]{#1\lvert#2 #1\rvert}
\newcommand{\norm}[2][]{#1\lVert#2 #1\rVert}

\newcommand{\tuple}[2][]{#1 \langle #2 #1 \rangle}
\newcommand{\ip}[3][]{\tuple[#1]{#2,#3}}

\renewcommand{\defas}{\overset{\text{\smash{\tiny{def}}}}{=}}

\newcommand\optparen[1]{\ifthenelse{\equal{#1}{}}{}{(#1)}}
\newcommand{\RiskChar}{L}
\newcommand{\Risk}[2]{\RiskChar_{#1}\protect\optparen{#2}}
\newcommand{\EmpRisk}[2]{\RiskChar_{#1}\protect\optparen{#2}}
\newcommand{\BinRisk}[2]{\RiskChar^{0-1}_{#1}\protect\optparen{#2}}

\newcommand{\rnd}[2]{\frac{\dee #1}{\dee #2}}

\newcommand{\ks}{\alpha}

\newcommand{\HS}{\mathcal H}

\algnewcommand{\LineComment}[1]{\State \(\triangleright\) #1}

\newcommand{\err}{\mathcal{E}}
\newcommand{\MI}[3][]{I_{#1}(#2;#3)}
\newcommand{\cMI}[4][]{I_{#1}(#2;#3|#4)}
\newcommand{\dcMI}[5][]{I^{#2}_{#1}(#3;#4|#5)}
\newcommand{\KLterm}{$m^{-1}\KL{Q}{P}$}
\newcommand{\oR}{\overline R}

\usepackage[T1]{fontenc}
\usepackage{inconsolata}
\usepackage{times}

\newcommand{\momentbound}{moment bound}
\newcommand{\Pinskerbound}{Pinsker bound}
\newcommand{\doprior}{data-oracle priors}

\usepackage{listings} %

\usepackage{enumitem}
\usepackage[utf8]{inputenc} %
\usepackage{booktabs}       %
\usepackage{amsfonts}       %
\usepackage{nicefrac}       %
\usepackage{enumitem}
\usepackage{siunitx}

\usepackage[hide=true,setmargin=true,marginparwidth=.8in]{marginalia}

\renewcommand{\algorithmicrequire}{\textbf{Hyperparameters:}}
\renewcommand{\algorithmicensure}{\textbf{Given:}}

\newcommand*\Let[2]{\State #1 $\gets$ #2}
\algrenewcommand\algorithmicrequire{\textbf{Given:}}
\algrenewcommand\algorithmicensure{\textbf{Hyperparameters:}}

\makeatletter
\renewcommand\paragraph{\@startsection{paragraph}{4}{\z@}%
                                    {1.25ex \@plus1ex \@minus.2ex}%
                                    {-1em}%
                                    {\normalfont\normalsize\bfseries}}
\makeatother
\usepackage{caption}

\usepackage[colorlinks,citecolor=blue,urlcolor=blue,linkcolor=RawSienna]{hyperref}
\usepackage[capitalize]{cleveref}

\crefname{lemma}{Lemma}{Lemmas}
\crefname{corollary}{Corollary}{Corollaries}
\crefname{theorem}{Theorem}{Theorems}
\crefname{claim}{Claim}{Claims}

\begin{document}

\title{On the role of data in PAC-Bayes bounds}

\author{Gintare Karolina Dziugaite$^1$}
\address{$^1$Element AI, $^2$University of Toronto, $^3$Vector Institute}

\author{Kyle Hsu$^{2,3}$}
\author{Waseem Gharbieh$^1$}

\author{Gabriel Aprino$^{2}$}

\author{Daniel M. Roy$^{2,3}$}

\begin{abstract}
The dominant term in PAC-Bayes bounds is often the Kullback--Leibler divergence between the posterior and prior.
For so-called linear PAC-Bayes risk bounds based on the empirical risk of a fixed posterior kernel, 
it is possible to minimize the expected value of the bound 
by choosing the prior to be the expected posterior, 
which we call the \emph{oracle} prior on the account that it is distribution dependent.
In this work, we show that the bound based on the oracle prior can be suboptimal:
In some cases, a stronger bound is obtained by using a data-dependent oracle prior, i.e., a
conditional expectation of the posterior, given a subset of the training data that is then excluded from the empirical risk term.
While using data to learn a prior is a known heuristic, its essential role in optimal bounds is new.
In fact, we show that using data can mean the difference between vacuous and nonvacuous bounds.
We apply this new principle in the setting of nonconvex learning,
simulating data-dependent oracle priors on MNIST and Fashion MNIST with and without held-out data,
and demonstrating new nonvacuous bounds in both cases.
\end{abstract}

\maketitle

\section{INTRODUCTION}

In this work, we are interested in the application of 
PAC-Bayes bounds \citep{McAllester1999,shawe1997pac} 
to the problem of understanding the generalization properties of learning algorithms.
Our focus will be on supervised learning from i.i.d.\ data,
although PAC-Bayes theory has been generalized far beyond this setting,
as summarized in a recent survey by \citet{Guedjsurvey}.
In our setting, 
PAC-Bayes bounds control the risk
of Gibbs classifiers,
i.e., randomized classifiers whose predictions, on each input, are determined by a classifier $h$ sampled according to some distribution $Q$ on the hypothesis space $\HS$.
The hallmark of a PAC-Bayes bound is a normalized Kullback--Leibler (KL) divergence, \KLterm{},
defined in terms of a Gibbs classifier $P$ that is called a ``prior'' because 
it must be independent of the $m$ data points used to estimate the empirical risk of $Q$.

In applications of PAC-Bayes bounds to generalization error, 
the contribution of the KL divergence often dominates the bound:
In order to have a small KL with a strongly data-dependent posterior, 
the prior must, in essence, predict the posterior.
This is difficult without knowledge of (or access to) the data distribution,
and represents a significant statistical barrier to achieving tight bounds.
Instead, 
many PAC-Bayesian analyses rely on generic priors chosen for analytical convenience.

Generic priors, however, are not inherent to the PAC-Bayes framework:
every valid prior yields a valid bound.
Therefore, if
one does not optimize the prior to the data distribution,
one may obtain a bound that is loose on the account of 
ignoring important, favorable properties of the data distribution.

\citet{langford2003microchoice} were the first to consider the problem of optimizing the prior to 
minimize the \emph{expected value} of the high-probability PAC-Bayes bound.
In the realizable case, they show that the problem reduces to optimizing the expected value of the KL term.
More precisely, they consider a fixed learning rule $S \mapsto Q(S)$, i.e., a fixed posterior kernel, 
which chooses a posterior, $Q(S)$, based on a training sample, $S$.
In the realizable case, the bound depends linearly on the KL term.
Then $\EE [\KL{Q(S)}{P}]$ is minimized by the expected posterior, $P^* = \EE[Q(S)]$,
i.e., $P^*(B) = \EE[Q(S)(B)]$ for measurable $B \subseteq \HS$.
Both expectations are taken over the unknown distribution of the training sample, $S$.
We call $P^*$ the \emph{oracle} prior.
If we introduce an $\HS$-valued random variable $H$ satisfying $\Pr[H|S] = Q(S)$ a.s.,
we see that its distribution, $\Pr[H]$, is $P^*$ and thus, the ``optimality'' of the oracle $P^*$ 
is an immediate consequence of the identity
$
\textstyle \MI{S}{H} = \EE[\KL{Q(S)}{P^*}] = \inf_{P'} \, \EE[\KL{Q(S)}{P'}],
$
a well-known variational characterization of mutual information in terms of KL divergence.

For so-called linear PAC-Bayes bounds (introduced below),
the oracle prior is seen to minimize the bound in expectation when all the data are used to estimate the risk. This holds even in the unrealizable setting.
In light of this, having settled on a learning rule $S \mapsto Q(S)$,
we might seek to achieve the tightest linear PAC-Bayes bound in expectation by
attempting to approximate the oracle prior, $P^*$.
Indeed, there is a large literature aimed at obtaining localized PAC-Bayes bounds via 
distribution-dependent priors, whether analytically \citep{Catoni,lever2010distribution},
through data \citep{Amb07,negrea2019information}, 
or by way of concentration of measure, privacy, or stability \citep{Oneto2016,Oneto2017,DR18,rivasplata2018pac}.

One of the contributions of this paper is the demonstration that
an oracle prior may not yield the tightest linear PAC-Bayes risk bound in expectation,
\emph{if we allow ourselves to consider also using only subsets of the data to estimate the risk}. 
\cref{mainthm} gives conditions on a learning rule for there to exist 
data-dependent priors that improves the bound based upon the oracle prior.
This phenomenon is a hitherto unstated principle of PAC-Bayesian analysis: 
data-dependent priors are sometimes necessary for tight bounds.
Note that, as the prior must be independent of data used to compute the bound \emph{a posteriori},
if $m$ training data are used to define the prior, only the remaining $n-m$ data should be used to compute the bound
(i.e., compute the empirical risk term and divide the KL term).
\emph{Note that  all $n$ training data are used by the learning algorithm.} 
We formalize these subtleties in the body of the paper and discuss some other misconceptions in \cref{app:misconceptions}.

We give an example of a learning problem where \cref{mainthm} implies data-dependent priors dominate. The example is adapted from a simple model of SGD in a linear model by \citet{NagaKolter19c}.
In the example, most input dimensions are noise with no signal and this noise accumulates in the learned weights. 
In our version, we introduce a learning rate schedule, and so earlier data points have a larger influence on the resulting weights.
Even so, there is enough variability in the posterior that the oracle prior yields a vacuous bound.
By conditioning on early data points, we reduce the variability and obtain nonvacuous bounds.

The idea of using data-dependent priors to obtain tighter bounds is not new \citep{Amb07,parrado2012pac,DR18,rivasplata2018pac}.
The idea is also implicit in the luckiness framework \citep{shawe1996framework}.
However, the observation that using data can be essential to obtaining a tight bound, even in full knowledge of the true distribution,
is new, and brings
a new dimension to the problem of constructing data-dependent priors.

In addition to demonstrating the theoretical role of data-dependent priors,
we investigate them empirically,
by studying generalization in nonconvex learning by stochastic (sub)gradient methods.
As data-dependent oracle priors depend on the unknown distribution,
we propose to use held-out data (``ghost sample'') to estimate unknown quantities.
Unlike standard held-out test set bounds, this approach relies implicitly on a type of stability demonstrated by SGD.
We also propose approximations to data-dependent oracle priors that use no ghost sample,
and find, given enough data, the advantage of the ghost sample diminishes significantly.
We show that both approaches yield state-of-the-art nonvacuous bounds on MNIST and Fashion-MNIST
for posterior Gaussian distributions whose means are clamped to the weights learned by SGD. Our MNIST bound (11\%) improves significantly on the best published bound (46\%) \citep{Zhou18}.
\NA{Finally, we evaluate minimizing a PAC-Bayes bound with our data-dependent priors as a learning algorithm.
We demonstrate significant improvements to both classifier accuracy and bound tightness, 
compared to optimizing with generic priors.}

\section{PRELIMINARIES}

Let $Z$ be a space of labeled examples, and write $\ProbMeasures{Z}$ for the space of (probability) distributions on $Z$.
Given a space $\HS$ of \defn{classifiers} (e.g., neural network predictors defined by their weights $w$) and a bounded \defn{loss function} $\loss: \HS \times Z \to [0,1]$,
the risk of a hypothesis $w \in \HS$ is
$
\Risk{\Dist}{w} = \EE_{z \sim \Dist} [\loss(w,z)].
$
We also consider \defn{Gibbs classifiers}, i.e., elements $P$ in the space $\ProbMeasures{\HS}$ of 
distributions on $\HS$,
where risk is defined by $\Risk{\Dist}{P} = \EE_{w \sim P} \Risk{\Dist}{w}$.
As $\Dist$ is unknown, learning algorithms often work by optimizing an objective that depends on i.i.d.\ 
training data $S \sim \Dist^n$, such as the \defn{empirical risk}
$
\EmpRisk{S}{w} = \EmpRisk{\smash{\hat \Dist_n}}{w} = \frac 1 n \sum_{i=1}^{n} \loss(w,z_i),
$
where $\hat \Dist_n$ is the empirical distribution of $S$.
Writing $Q(S)$ for a data-dependent Gibbs classifier (i.e., a \emph{posterior}),
our primary focus is its risk, $\Risk{\Dist}{Q(S)}$, and its relationship to empirical estimates, such as $\EmpRisk{S}{Q(S)}$. 

The PAC-Bayes framework~\citep{McAllester1999,shawe1997pac} 
provides generalization bounds on  data-dependent Gibbs classifiers.
Let $Q,P \in \ProbMeasures{\HS}$ be probability measures defined on a common measurable space $\HS$.
When $Q$ is absolutely continuous with respect to $P$, written $Q \ll P$,
we write $\rnd{Q}{P} : \HS \to \NNReals \cup \{\infty\}$ for some Radon--Nikodym derivative (aka, density) of $Q$ with respect to $P$.
The Kullback--Liebler (KL) divergence
from $Q$ to $P$ is
$\KL{Q}{P} = \int \ln \rnd{Q}{P} \,\dee Q$ if $Q \ll P$ and $\infty$ otherwise.
Assuming $Q$ and $P$ admit densities $q$ and $p$, respectively, w.r.t.\ some sigma-finite measure $\nu \in \Measures{\HS}$,
the definition of the KL divergence satisfies
\begin{equation*}
\KL{Q}{P} = \int \log \frac{q(w)}{p(w)} q(w) \nu(\dee w).
\end{equation*}

The following PAC-Bayes bound follows from \citep[][Thm.~2]{McAllesterDropOut}, taking $\beta = 1 - 1/(2\lambda)$. (See also
\citet[][Thm.~1.2.6]{Catoni}.)

\newcommand{\LPBB}[5]{\Psi_{#1,#5}(#3,#4;#2)}
\newcommand{\klLPBB}[4]{\Psi^{*}_{#4}(#2,#3;#1)}
\newcommand{\diffLPBB}[4]{\Psi^{\dagger}_{#4}(#2,#3;#1)}
\begin{theorem}[Linear PAC-Bayes bound]
\label{pacbayeslinear}
Let $\beta,\delta \in (0,1)$,
$n \in \Nats$,
$\Dist \in \ProbMeasures{Z}$,
and
$P \in \ProbMeasures{\HS}$.
With probability at least $1-\delta$ over $S \sim \Dist^n$,
for all $Q \in \ProbMeasures{\HS}$,
\begin{equation*}%
 \Risk{\Dist}{Q} \le 
 \LPBB{\beta}{S}{Q}{P}{\delta}
 \defas
 \frac 1 \beta \EmpRisk{S}{Q}
	+ \frac { \KL{Q}{P} + \log \frac {1}{\delta} }{2 \beta(1-\beta) |S|}.
\end{equation*}
\end{theorem}

As is standard, we call $P$ the \defn{prior}.

Note that the KL term in the bound depends on the data $S$ through the kernel $Q(S)$.
If we are interested in obtaining the tightest possible bound for the kernel $Q(S)$, 
then we can seek to minimize the KL term in some distribution sense.
Our control of the KL term comes from the prior $P$.
Since the bound is valid for all priors independent from $S$,
we can choose $P$ by optimizing, e.g., the risk bound in expectation, as first proposed by \citet{langford2003microchoice}:

\begin{theorem}%
\label{oracleprior}
Let $n \in \Nats$ and fix a probability kernel $Q : Z^n \to \ProbMeasures{\HS}$.
For all $\beta,\delta \in (0,1)$ 
and $\Dist \in \ProbMeasures{Z}$,
$\EE_{S \sim \Dist^n} \LPBB{\beta}{S}{Q(S)}{P}{\delta}$
is minimized, in $P$, 
by the ``oracle'' prior $P^* = \EE_{S \sim \Dist^n}[Q(S)]$.
\end{theorem}

Note that, in other PAC-Bayes bounds, the KL term sometimes appears within a concave function. In this case, 
oracle priors can be viewed as minimizing an upper bound on bound. We focus on linear PAC-Bayes bounds here for analytical tractability.

\newcommand{\vx}{\boldsymbol{x}}
\newcommand{\vu}{\boldsymbol{u}}
\newcommand{\vw}{\boldsymbol{w}}
\newcommand{\aeta}{\bar\eta}
\newcommand{\WW}{W}
\newcommand{\etas}[2]{\eta_{#1}^{#2}}

\section{DATA-DEPENDENT ORACLE PRIORS}
\label{mainexample}
\newcommand{\NIG}[4][]{\mathrm{R}_{#1}(#2;#3|#4)}
\newcommand{\ExB}[3]{\mathrm{B}(#1;#2|#3)}

\newcommand{\PF}{\mathcal F}

Here we demonstrate that, 
for linear PAC-Bayes bounds,
one may obtain a stronger bound using a ``data-dependent oracle'' prior, rather than the usual (data-independent) oracle prior. 
Further, using a data-dependent oracle prior may mean the difference between a vacuous and nonvacuous bound.

A typical PAC-Bayes generalization bound for a posterior kernel $S \mapsto Q(S)$ is based on the empirical risk
$\EmpRisk{S}{Q(S)}$ computed from the same data fed to the kernel.
Instead, let $J$ be a (possibly random) subset of $[n]$ of size $m<n$, independent from $S$,
let $S_J$ denote the subsequence of data with indices in $J$, and 
let $S \setminus S_J$ denote the complementary subsequence.
Consider now the PAC-Bayes bound based on the estimate
$\EmpRisk{S \setminus S_J }{Q(S)}$.
In this case, the prior need only be independent from $S \setminus S_J$.
The \defn{$\sigma(S_J)$-measurable data-dependent oracle prior $P^*(S_J) = \EE[Q(S)|S_J]$} arises
as the solution of the optimization \fTBD{DR: We may generalize this to consider a constrained optimization, restricting the kernel to taking values in, say, the family of Gaussian distributions. This would then allow us to formalize the actual derivations we make in our empirical study. If the family of (say, Gaussian) distributions were denoted $\cF$, we're considering denoting this optimum as $I^\cF(\hat h; S|S_J)$, as it would be literally the mutual information if the constraints were lifted.}
\[\label{condminfvar}%
\inf_{P \in Z^{|J|} \to \ProbMeasures{\HS}} \EE[ \KL{Q(S)}{P(S_J)} ].
\]
Letting $\hat w$ be a random element in $\HS$ satisfying $\Pr[\hat w | S,J] = Q(S)$ a.s.,
the value of \cref{condminfvar} is the \emph{conditional} mutual information $\cMI{\hat w}{S}{S_J}$. 
This conditional mutual information represents the expected value of the KL term in the linear PAC-Bayes bound and 
so this data-dependent prior achieves, in expectation, the tightest linear PAC-Bayes bound based on the estimate
$\EmpRisk{S \setminus S_J }{Q(S)}$.

We can
also consider restricting the prior distribution to a family $\PF \subseteq \ProbMeasures{\HS}$ of distributions,
in which case the optimization in \cref{condminfvar} is over the set of kernels $Z^{|J|} \to \PF$.
We refer to a solution of this optimization as a data-dependent oracle prior \emph{in $\PF$},
denoted $P^*_{\PF}(S_J)$,
and refer to the value of \cref{condminfvar} as the conditional $\PF$-mutual information, denoted
$\cMI[\PF]{\hat w}{S}{S_J}$. The unconditional $\PF$-mutual information is defined equivalently.\footnote{%
When $\PF$ is the set of all distributions, we drop $\PF$ from the notation. The notation $P^*(S_J)$ is understood to also specify
the data $S_J$ held out from the estimate of risk. Thus, $P^*_{\PF}$ denotes the distribution-dependent but data-independent oracle prior when the choice of prior is restricted to $\PF$, just as $P^*$ represents the distribution-dependent but data-independent oracle prior when the choice of prior is unrestricted.
}
In \cref{sec:sgdddpriors}, we study data-dependent oracle priors in a restricted family $\PF$ in a setting where dealing with the set of all priors is intractable.

Fix $\PF$ and define the \defn{information  rate gain} (from using $S_J$ to choose the prior in $\PF$)
and the \defn{excess bias} (from using $S \setminus S_J$ to estimate the risk) to be, respectively,
\[
\NIG[\PF]{\hat w}{S}{S_J} = 
\frac { \MI[\PF]{\hat{h}}{S}}{|S|} - \frac { \cMI[\PF]{\hat{h}}{S}{S_J,J} }{|S\setminus S_J|}
\]
and
\[
\ExB{\hat w}{S}{S_J} = 
\EE[ \EmpRisk{S \setminus S_J}{\hat w} - \EmpRisk{S}{\hat w}].
\]
Note that, if $J$ is chosen uniformly at random, then $\ExB{\hat w}{S}{S_J}=0$.
Using these two quantities, we can characterize whether a data-dependent prior can outperform the oracle prior.
The following result is an immediate consequence of the above definitions. 
(We present the straightforward proof in \cref{app:mainthm} for completeness.)

\begin{proposition}\label{mainthm}
Let $\beta,\delta \in (0,1)$,
$n \in \Nats$, and
$\Dist \in \ProbMeasures{Z}$.
Fix $Q : Z^n \to \ProbMeasures{\HS}$
and let $J \subseteq [n]$ be a (possibly random) subset of nonrandom cardinality $m<n$, independent from $S \dist \Dist^n$. 
Conditional on $S$ and $J$, let $\hat w$ have distribution $Q(S)$.
Then
$\EE_J \EE_{S \sim \Dist^n}\LPBB{\beta}{S \setminus S_J}{Q(S)}{P^*_{\PF}(S_J)}{\delta} 
    \le \EE_{S \sim \Dist^n} \LPBB{\beta}{S}{Q(S)}{P^*_{\PF}}{\delta}$ 
    if and only if
\begin{equation}\label{keyinequality}
\textstyle
\NIG[\PF]{\hat w}{S}{S_J} 
\ge
2(1-\beta)\, \ExB{\hat w}{S}{S_J} 
+ \frac{ \log \frac {1}{\delta} }{n} 
\frac {m}{n-m }
,
\end{equation}
i.e., \cref{keyinequality} holds if and only if the
linear PAC-Bayes bound with a oracle (data-independent) prior
is no tighter, in expectation, than that with the data-dependent oracle prior.
\end{proposition}

To interpret the proposition, consider $\beta = 1/2$:
then a data-dependent prior yields a tighter bound,
if the information rate gain is larger than
the excess bias and a term that accounts for excess variance.

Do such situations arise naturally?
In fact, they do. The following demonstration uses a linear classification problem presented by \citet{NagaKolter19c}.
Their example was originally constructed to demonstrate potential 
roadblocks to studying generalization in SGD using uniform convergence arguments.
We make one, but important modification: 
we modify the learning algorithm to have another feature of SGD in practice: a \emph{decreasing} step size.
As is the case in ordinary training, 
the decreasing step size
causes earlier data points to have more influence. 
As the data are noisy, the noise coming from these early samples has an outsized effect that renders a linear PAC-Bayes bound vacuous.
By leaving the initial data out of the estimate of risk, and using a data-dependent oracle prior,
we achieve a tighter bound.
Indeed, we obtain a nonvacuous bound, while the optimal data-independent oracle prior yields a \emph{vacuous} bound.

\begin{figure}
\includegraphics[width=.65\linewidth]{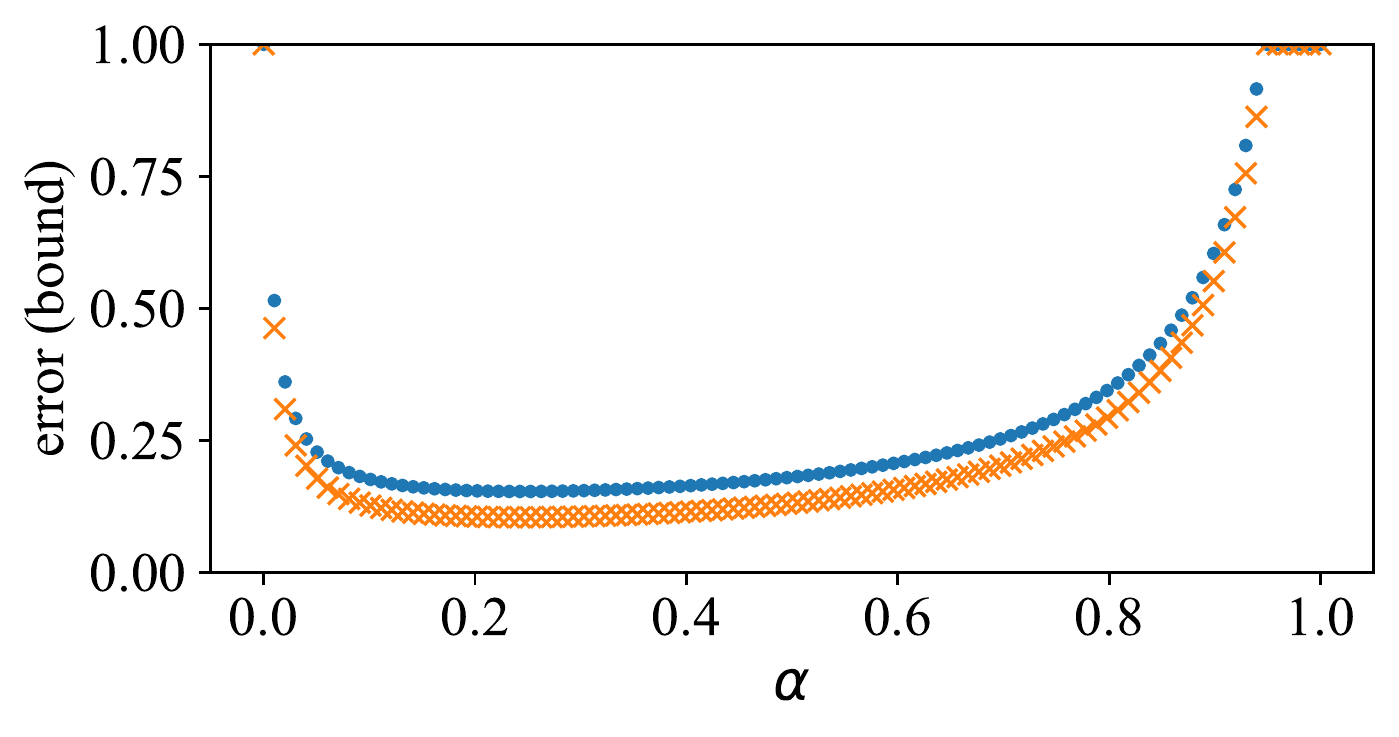}
\caption{Lower (orange x's) and upper (blue dots) bounds on the expected value of a linear PAC-Bayes bound
as a function of the fraction, $\alpha$, of the 100 training data used by the data-dependent (PAC-Bayes) prior.
Each bound uses the optimal (in expectation) tradeoff $\beta$ and data-dependent prior $P(S_J)$, for $J=[k]$.
Without using data (i.e., $J=\emptyset$), the bound is provably vacuous as the lower bound exceeds one.
The upper bound is approximately $0.15$ when the oracle prior is computed conditionally given the first 24 data points (i.e., $J=[24]$ and $\alpha=0.24$.).
}\label{fig:theorybound}

\end{figure}

\begin{example}\label{mainexample}
Consider the hypothesis class $\HS = \Reals^d$, interpreted as linear classifiers
\[
\vx \mapsto \sign(\ip{\vx}{\vw}) : \Reals^d \to \set{-1,0,1}, \quad \text{for $\vw \in \Reals^d$.}
\]
Assume that $d = K + D$, with $D \gg K$, and decompose each input $\vx \in \Reals^d$
as $\vx = (\vx_1,\vx_2)$, where $\vx_1 \in \Reals^K$ and $\vx_2 \in \Reals^D$. (We will decompose the weights similarly.)
Labels $y$ take values in $\set{\pm 1}$ and so a prediction of $0$ (i.e., on the decision boundary) is a mistake.

Consider the following $n$ i.i.d.\ training data:
Let $\vu \in \Reals^k$ be a nonrandom vector and, for each $i = 1,\dots,n$,
choose $y_{i}$ uniformly at random in $\set{\pm 1}$,
let $\vx_{i,1} = y_i \vu$,
and let $\vx_{i,2}$ be multivariate normal with mean 0 and covariance $(\sigma^2/D) \, I_D$,
where $I_D$ is the $D \times D$ identity matrix.
Let $\Dist$ denote the common marginal distribution of each training example $(y_{i},\vx_{i})$.

Consider the following one-pass learning algorithm:
Let $\vw_{0} = 0$, then, for $t=1,\dots,n$ and
$\eta_{t} = 1/t$,
put
$
\vw_{t} = \vw_{t-1} + \eta_{t} y_{t} \vx_{t}.
$
Then define the final weights to be $\WW = \vw_n + (0,\xi)$, where $\xi$ is an independent,
zero-mean multivariate Gaussian with covariance $\kappa\, I_D$.
Note that $\vw_{n} = (\vw_{n,1},\vw_{n,2})$ where
$\vw_{n,1} = (\sum_{i=1}^{n} \eta_{i}) \vu$
and $\vw_{n,2} = \sum_{i=1}^{n} \eta_{i} y_{i} \vx_{i,2}$.

We will compare bounds based on oracle priors with those based on data-dependent oracle priors.
To that end, let $S = \{{(y_i,\vx_i)}\}_{i=1}^{n}$
and define $Q$ by $\Pr[W|S] = Q(S)$ a.s.
Let $[n] = \{{1,\dots,n}\}$. 
For a subset $J \subseteq [n]$, let $S_J$ be the corresponding subset of the data $S$
and let $S\setminus S_J$ be the complement.

\begin{lemma} \label{keyclaim}
There are constants $n,D,\sigma,\kappa,\delta,u$
such that the infimum
\[\label{theinf}
\inf_{J,\beta, P}\,
\EE \big [ \LPBB{\beta}{S \setminus S_J}{Q(S)}{P(S_J)}{\delta} \big ],
\]
where $J$ ranges over subsets of $[n]$,
$\beta$ ranges over $(0,1)$,
and $P$ ranges over measurable functions $Z^{|J|} \to \ProbMeasures{\HS}$,
is achieved by a nonempty set $J$.
In particular, the optimal prior is data dependent.
\end{lemma}

Lower and upper bounds on the objective (\cref{theinf}) for $J$ of the form $\{1,\dots,\floor{100 \alpha}\}$, for $\alpha \in [0,1]$, are visualized in \cref{fig:theorybound}. Using a data-dependent prior in this scenario is critical for obtaining a nonvacuous bound. 
The derivation of these bounds as well as a sketch of the proof and a complete rigorous proof, can be found in \cref{app:fullproof}.
\end{example}

In summary, data-dependent oracle priors, by definition, minimize linear PAC-Bayes bounds in expectation. The example above demonstrates that data-dependence can be essential in using linear PAC-Bayes bounds to obtain nonvacuous bounds. The example relies in a crucial way on the step size decreasing, so that some data points have an outsized impact on the noise that is injected into the classifier. In the remainder, we consider the problem of exploiting data dependent priors in the setting of learning with SGD.

\newcommand{\full}{\emph{base}}
\newcommand{\prefix}{$\alpha$-\emph{prefix}}
\newcommand{\prefixghost}{$\alpha$-\emph{prefix+ghost}}

\section{DATA-DEPENDENT PRIORS FOR SGD}
\label{sec:sgdddpriors}

As the theoretical results in the previous section demonstrate,
data-dependent oracle priors can lead to dramatically tighter bounds.
In this section, we take the first steps towards understanding whether data-dependent priors can aid us in the study of deep learning with stochastic gradient descent (SGD). 

Most attempts to build nonvacuous PAC-Bayes bounds for neural networks learned by SGD fail when the bounds are derandomized \citep{neyshabur2017pac,Naga19}. 
In order to gain tight control on the derandomization, one requires that the posterior is concentrated tightly around the weights learned by SGD.
This leads to a significant challenge as the prior must accurately predict the posterior, otherwise the KL term explodes.
Can data-dependent priors allow us to use more concentrated priors? 
While we may not be able to achieve derandomized bounds yet,
 we should be able to build tighter bounds for stochastic neural networks with lower empirical risk.

In \cref{mainexample}, we studied a posterior that depended more heavily on some data points than others.
This property was introduced intentionally in order to serve as a toy model for SGD.
Unlike the toy model, however, we know of no representations of the marginal distribution of the parameters learned by SGD that would allow us to optimize or compute a PAC-Bayes bound with respect to a data-dependent oracle prior.
As a result, we are forced to make approximations.

Issues of tractability aside, 
another obstacle to using a data-dependent oracle prior is its dependence on the unknown data distribution.
Ostensibly, this statistical barrier can be surmounted with extra data,
although this would not make sense in a standard model-selection or self-bounded learning setup.
In these more traditional learning scenarios, one has a training data set $S$ and
wants to exploit this data set to the maximum extent possible.
Using some of this data to estimate or approximate (functionals of) the unknown distribution
means that this data is not available to the learning algorithm or the PAC-Bayes bound.
Indeed, if our goal is simply to obtain the tightest possible bound on the risk of our classifier,
we ought to use most of this extra data to learn a better classifier, leaving out a small fraction to get a tight Hoeffding-style estimate of our risk.

However, if our goal is to understand the generalization properties of some posterior kernel $Q$ (and indirectly an algorithm like SGD),
we do not simply want a tight estimate of risk.
\emph{Indeed, a held-out test set bound is useless for understanding as it merely certifies that a learned classifier generalizes.} 
If a classifier generalizes due to favorable properties of the data distribution,
then we must necessarily capture these properties in our bound. 
These properties may be 
natural side products of the learning algorithm (such as weight norms)
or functionals of the unknown distribution that we must estimate (such as data-dependent oracle priors or functionals thereof).
In this case, it makes sense to exploit held out data to gain insight.

\newcommand{\wwf}{w_S}
\newcommand{\wwa}{w_{\alpha}}
\newcommand{\wwaghost}{w^G_{\alpha}}
\newcommand{\covf}{\Sigma}
\newcommand{\cova}{\Sigma_{\alpha}}
\newcommand{\Sa}{S_{\alpha}}

\newcommand{\gdata}{S^G}
\newcommand{\gdataa}{S^G_{\alpha}}
\newcommand{\pvar}{\sigma_{P}}

\newcommand{\cPr}[2]{\Pr^{#1}[#2]}
\newcommand{\cEE}[2]{\EE^{#1}[#2]}

\newcommand{\ddprior}{P}
\newcommand{\post}{Q}

\newcommand{\RS}{U}
\subsection{Optimal isotropic Gaussian priors}

In order to make progress, 
we begin by optimizing a prior over a restricted family $\PF$.
In particular, we consider the family of Gaussian priors when the posterior kernel chooses Gaussian posteriors.
Based on empirical findings on the behavior of SGD in the literature,
we propose an approximation to the data-dependent oracle prior.

Let $(\Omega,\mathcal F,\nu)$ be a probability space representing the distribution of a source of randomness.
Our focus here is on kernels $Q : \Omega \times Z^n  \to \ProbMeasures{\HS}$
where $Q(\RS, S) = \Normal{\wwf}{\covf}$ is a multivariate normal,
centered at the weights $w_S \in \Reals^{\pdim}$ learned by SGD (using randomness $U$, which we may assume without loss of generality encodes both the random initialization and the sequence of minibatches) on the full data set, $S$.
Such posteriors underlie several recent approaches to obtaining PAC-Bayes bounds for SGD.
In these bounds,
the covariance matrix $\covf$ is chosen to be diagonal and the scales are chosen to allow one to derive the bound on a deterministic classifier from the bound on a randomized classifier $Q$.
For example, \citet{neyshabur2017pac} derive deterministic classifier bounds from a PAC-Bayes bound based on (an estimate of) the Lipschitz constant of the network.

Fix some nonnegative integer $m\le n$ and let $\alpha = m/n$. Let $\Sa$ denote the size $m$ subset of $S$
corresponding to the first $m$ indices processed by SGD.
(Note that these indices are encoded in $U$.)
Writing $\cEE{\Sa,U}{\cdot}$ for the conditional expectation operator given $\Sa,U$,
\cref{oracleprior} implies that
the tightest (linear PAC-Bayes) bound in expectation is obtained by
minimizing $\cEE{\Sa,\RS}{\KL{\post(\RS,S)}{\ddprior}}$ in terms of $P$, 
which yields the data-dependent oracle prior $\ddprior = \cEE{\Sa,\RS}{\post(\RS,S)}$.
(We are permitted to condition on $U$ because $U$ is independent from $S$.)

As this prior is assumed to be intractable and the data distribution is unknown, we make a few approximations.
First, as proposed in \cref{mainexample},
we consider optimizing the prior over a family $\PF$ of priors. 
Specifically, consider the identifying the isotropic Gaussian prior
$\ddprior = \Normal{\wwa}{\pvar I}$
that minimizes $\cEE{\Sa,\RS}{\KL{Q(U,S)}{\ddprior}}$.
(We will revisit this simplification in \cref{sec:oraclevar}, where we consider priors and posteriors with non-isotropic diagonal covariance matrices.)
If we fix $\pvar$, then based on the KL divergence between multivariate Gaussians (\cref{klgaussians}),
the optimization problem reduces to 
\begin{equation}\label{gaussopt}
\arg\min_{\wwa} \,\cEE{\Sa,U}{\norm{\wwf - \wwa}^2 }.
\end{equation}
It follows that the mean of the \emph{Gaussian} oracle prior (with fixed isotropic covariance) 
is the conditional expectation $\cEE{\Sa,U}{\wwf}$ of the weights learned by SGD.
Under this choice, the contribution of the mean component to the bound is the value of the expectation in \cref{gaussopt},
which can be seen to be the trace of the conditional covariance of $\wwf$ given $\Sa,U$.
For the remainder of the section we will focus on the problem of approximating the oracle prior mean. 
The optimal choice of $\pvar$ depends on the distribution of $\covf$. One approach, which assumes that we build separate bounds for different values of $\pvar$ that we combine via a union bound argument, is outlined in \cref{app:approx}.

\subsection{Ghost samples}
\label{sec:coupling}

\begin{figure}[t]
\scalebox{.6}{
\includegraphics[width=.25\linewidth]{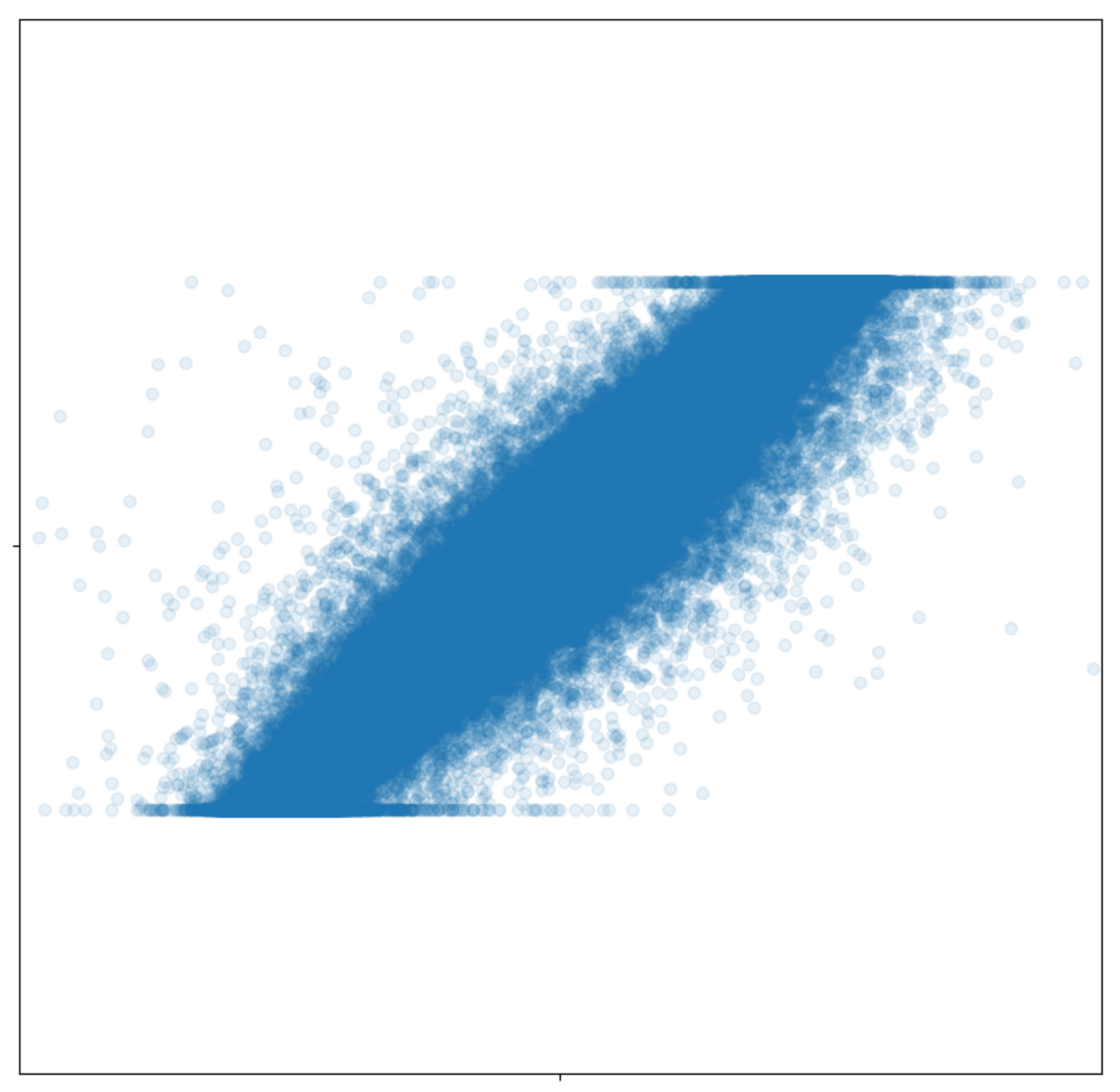}%
\includegraphics[width=.25\linewidth]{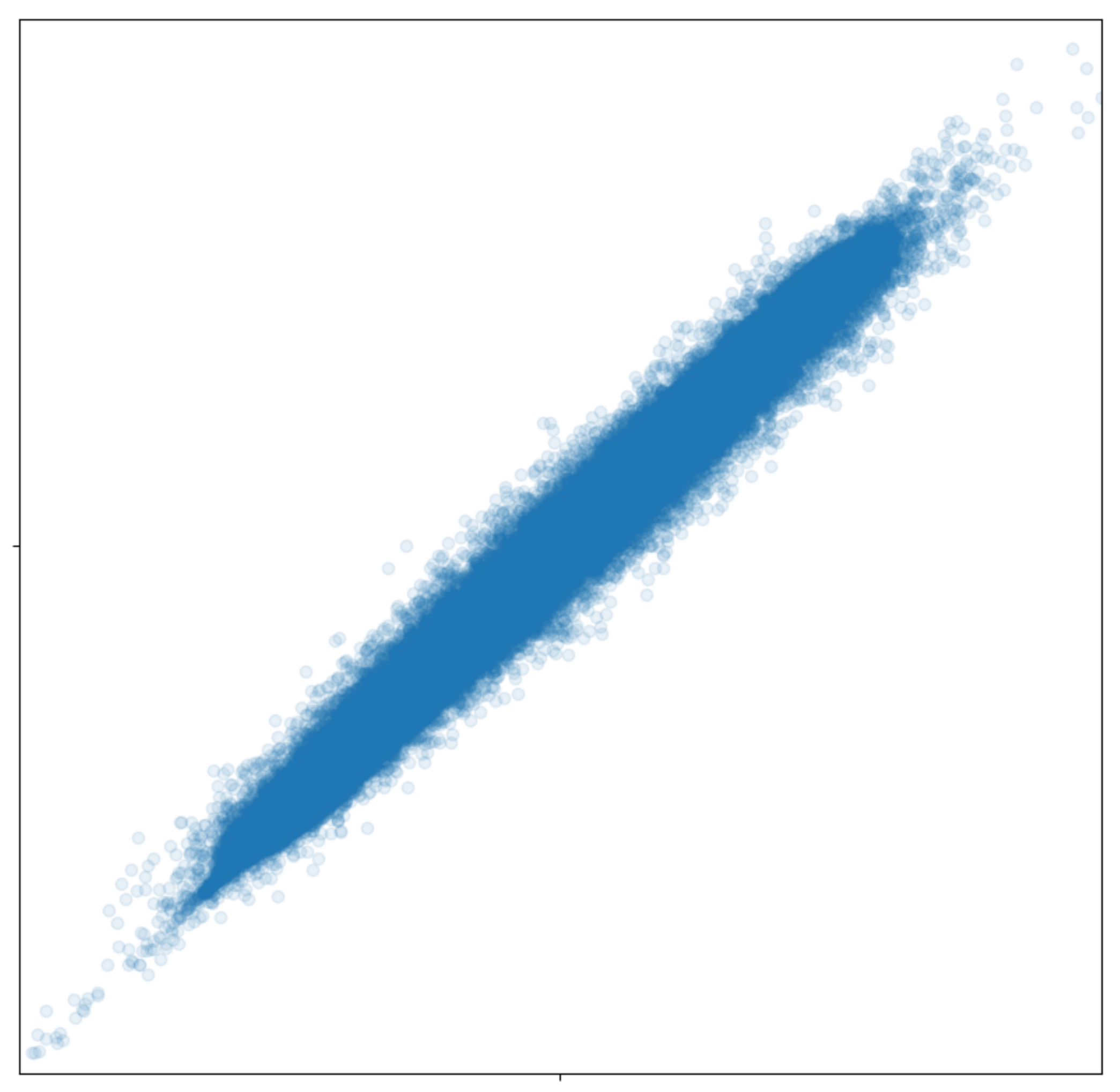}%
\includegraphics[width=.25\linewidth]{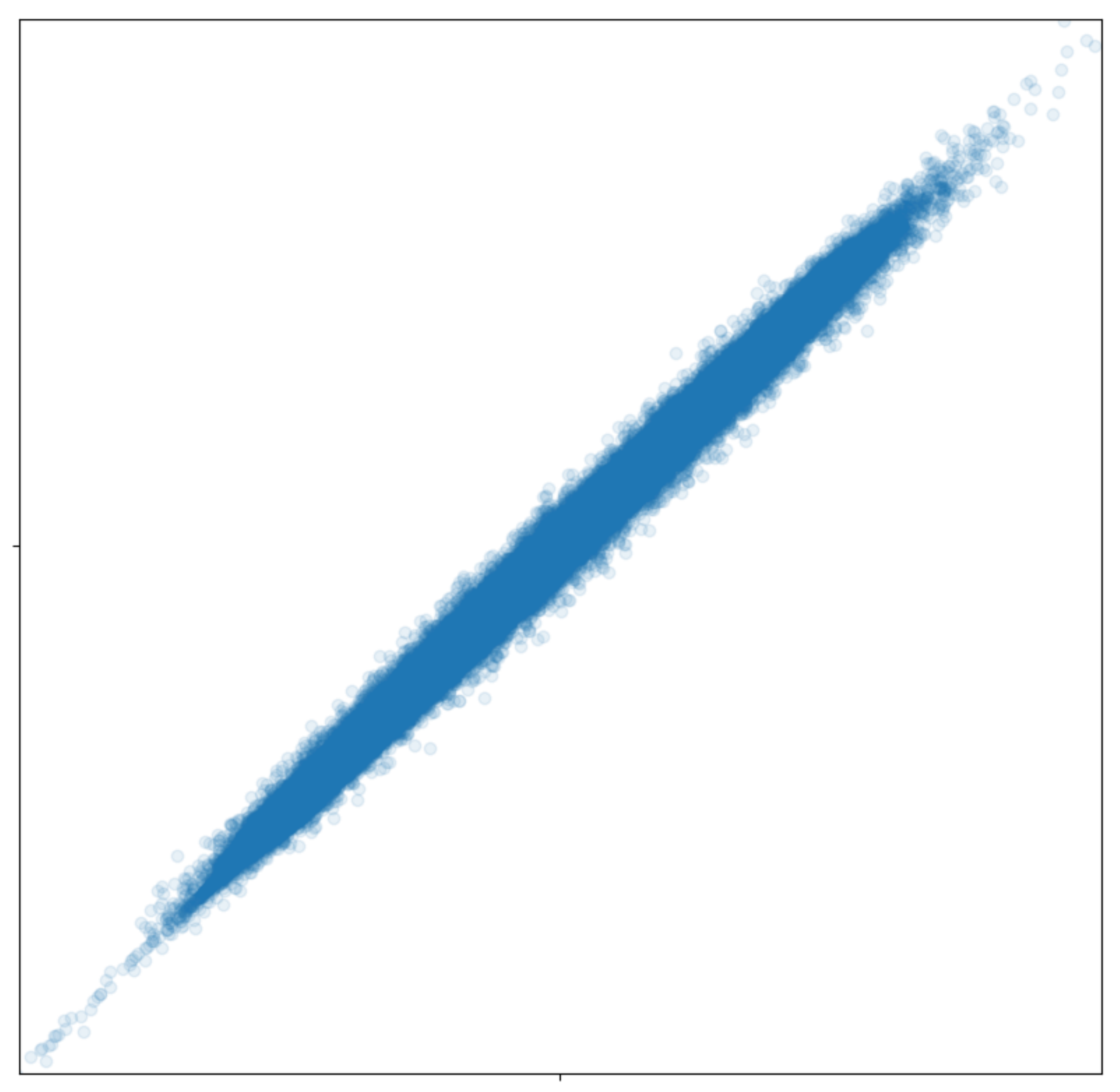}%
\includegraphics[width=.25\linewidth]{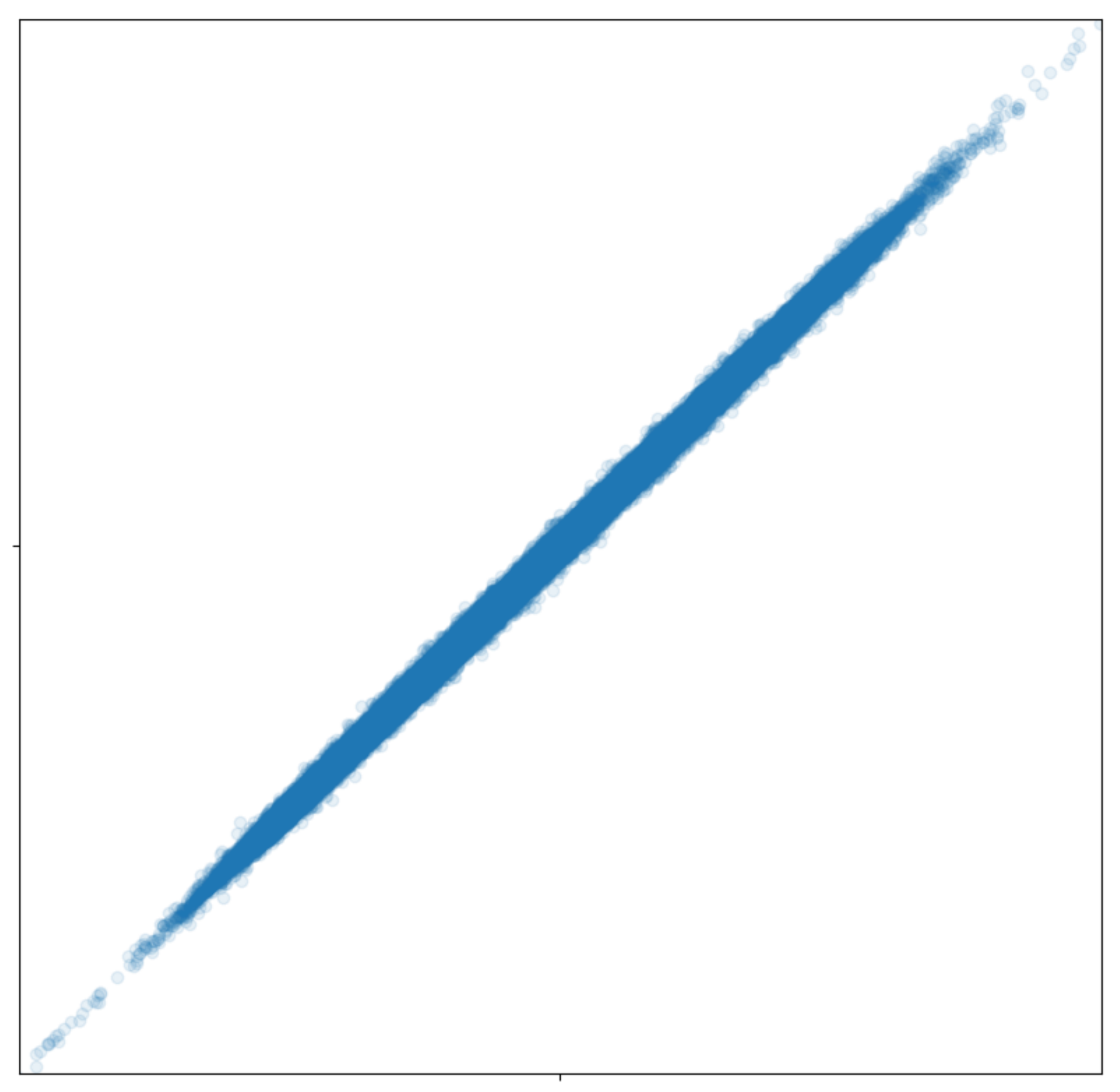}
}
\caption{
MNIST, FC;
\textbf{x-axis}: parameter values of \full{} run;
\textbf{y-axis}: parameter values of \prefix{} run;
\textbf{left to right:} $\alpha$ values equal to $\{0, 0.1, 0.5, 0.9 \} $.
As $\alpha$ increases, the correlation between the parameters learnt by SGD on all of the data and an $\alpha$ fraction of the data increases.
}
\label{fig:correlationplots}
\end{figure}

\begin{algorithm*}[t]
  \caption{PAC-Bayes bound computation (right) and optimization (left).
          \textbf{Given}:Data $S$, ghost data $\gdata$ (if \prefixghost{}), batch size $b$. \textbf{Hyperparameters}:stopping criteria $\err$, prefix fraction $\alpha$, prefix stopping time $T$, prior variance $\sigma_P$.
}
\label{mainalgs}
\begin{minipage}[t]{.54\linewidth}
    \begin{algorithmic}
        \Function{Bound-Opt }{$\alpha, \sigma_P, T, \eta$}
        \Let{$S_{\alpha}$}{$\{z_1,..,z_{\alpha |S|} \}  \subset S$} \Comment{Select \prefix{}}
        \Let{$w^0_{\alpha}$}{SGD($ w_0, S_{\alpha},b,\frac{|S_{\alpha}|}{b})$} \Comment{Coupling}
        \Let{$w_{\alpha}^S$}{SGD($w^0_{\alpha}, S,  b,\infty, 0$)} \Comment{\prefix{}}
        \Let{$P$}{$\Normal{w_{\alpha}^S}{\sigma_P I_{p}}$}
        \Let{$\theta_Q$}{$(w^0_{\alpha}, \sigma_P )$} \Comment{$Q$ trainable params}
        \LineComment{Let $Q(\theta_Q)  = \Normal{w^0_{\alpha}}{\sigma_P I_{p}}$}
        \For{$i \gets 1 \textrm{ to } T$}
      	    \State{Sample minibatch $S' \in S \setminus S_{\alpha}$, $|S'|=b$.}
            \Let{$\theta_Q$}{$\theta_Q - \eta \grad_{\theta_Q} \diffLPBB{S\setminus S_{\alpha}}{Q(\theta_Q)}{P}{\delta} $}
        \EndFor
        \Let{Bound}{ $\klLPBB{S\setminus S_{\alpha}}{Q(\theta_Q)}{P}{\delta}$}
        \State \Return{Bound}
        \EndFunction
    \end{algorithmic}%
\end{minipage}%
\hspace*{.02\linewidth}
\begin{minipage}[t]{.44\linewidth}
  \begin{algorithmic}
    \Function{Get-Bound}{$\err, \alpha, T, \sigma_P$}
    \Let{$S_{\alpha}$}{$\{z_1,..,z_{\alpha |S|} \}  \subset S$} %
      \Let{$w^0_{\alpha}$}{SGD($ w_0, \Sa,b,\frac{|S_{\alpha}|}{b}$)} %
      \LineComment{Perform \full{} run}
      \Let{$w_{S}$}{SGD($w^0_{\alpha}, S,  b,\infty, \err$)} 
      \LineComment{Perform \prefixghost{} run}
      \Let{$\wwaghost$}{SGD( $w^0_{\alpha}, \gdataa,b,T,\cdot$)} 
      \Let{$P$}{$\Normal{\wwaghost}{\sigma_P I_{p}}$}
      \Let{$Q$}{$\Normal{w_{S}}{\sigma_P I_{p}}$}

       \Let{Bound}{$\klLPBB{S\setminus S_{\alpha}}{Q}{P}{\delta}$}
      \State \Return{Bound}
    \EndFunction
  \end{algorithmic}
\end{minipage}

\end{algorithm*}

In the setting above, the optimal Gaussian prior mean is given by the conditional expectation $\cEE{\Sa,U}{\wwf}$.
Although the distribution $\Dist$ is presumed to be unknown, there
is a natural statistical estimate for $\cEE{\Sa,U}{\wwf}$. Namely,
consider a \defn{ghost sample}, $\gdata$, independent from and equal in distribution to $S$.
Let $\gdataa$ be the data set obtained by combining $\Sa$ with a $1-\alpha$ fraction of $\gdata$.
(We can do so by matching the position of $\Sa$ within $S$ and within $\gdataa$.)
Note that $\gdataa$ is also equal in distribution to $S$.
We may then take $\wwaghost$ to be the mean of $Q(U,\gdataa)$,
i.e., the weights produced by SGD on the data set $\gdataa$ using the randomness $U$.

By design, SGD acting on $\gdataa$ and randomness $U$ will process $\Sa$ first and then start processing the data from the ghost sample.
Crucially, the initial $\alpha$ fraction of the first epoch in both runs will be identical.
By design, $\wwaghost$ and $\wwf$ are equal in distribution when conditioned on $\Sa$ and $U$, 
and so $\wwaghost$ is an unbiased estimator for
$\cEE{\Sa,U}{\wwf}$.%
\footnote{We can minimize the variance of the KL term by producing conditionally i.i.d. copies of $\wwaghost$ and averaging, although each such copy requires an independent $n-m$-sized ghost sample.}

\subsection{Terminology}
\label{sec:term}
We call the run of SGD on data $\Sa$ the \prefix{} run. 
The run of SGD on the full data is called the \full{} run.
A prior is constructed from the \prefix{} run by centering a Gaussian 
at the parameters obtained after $T$ steps of optimization. %
Prefix stopping time $T$ is chosen from a discrete set of values to minimize $L^2$ distance to posterior mean.%
\footnote{We account for these data-dependent choices via a union bound, which produces a negligible contribution.}
Note, that for $\alpha = 0$, $w_{\alpha} = w_0$, i.e., the prior is centered at random initialization as it has no access to data. 
This is equivalent to the approach taken by \citet{DR17}. 
When the prior has access to data $\gdataa$, we call an SGD run training on $\gdataa$ an \prefixghost{} run, obtaining parameters $\wwaghost$.

\begin{figure*}[t]
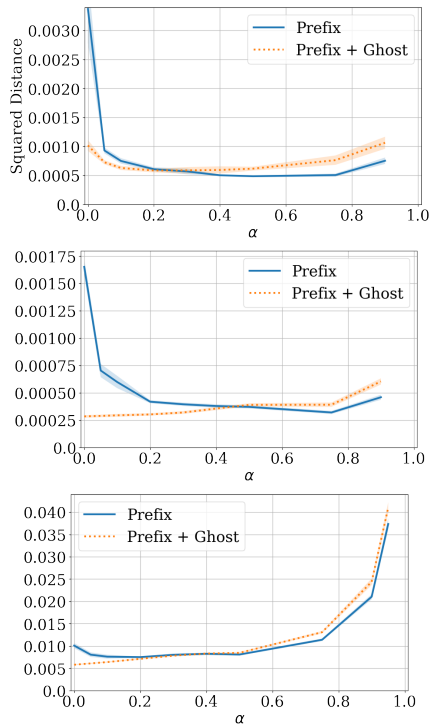

\centering 
\includegraphics[height=32mm]{Scaled_Min_L2_MLP_MNIST}\\%
\includegraphics[height=32mm]{Scaled_Zhou_MNIST}\\%
\includegraphics[height=32mm]{Scaled_Zhou_Fashion}
\caption{
\textbf{top}: MNIST, LeNet-5;
\textbf{center}: Fashion-MNIST, LeNet-5;
\textbf{bottom}: MNIST, FC;
\textbf{x-axis}: $\alpha$ used for \prefix{} \prefixghost{} runs;
\textbf{y-axis}: squared $L^2$ distance divided by $(1-\alpha) |S|$. For a Gaussian priors and posteriors with fixed covariance, smaller distances yields tighter bounds.
}
\label{fig:l2distanceplots}
\end{figure*}

The procedure of running the \prefix{} and \full{} runs together for the first $\alpha$-fraction of a \full{} run epoch using shared information $U$ (storing the data order) is an example of a \emph{coupling}. This coupling is simple and does not attempt to match  \full{} and \prefix{} runs beyond the first $m/b$ iterations (where $b$ is the batch size, which we presume divides $m$ evenly for simplicity).
It exploits the fact that the final weights have an outsized dependence on the first few iterations of SGD.
More advanced coupling methods can be constructed.
Such methods might attempt to couple beyond the first $\alpha$--fraction of the first epoch.

\begin{wrapfigure}{L}{0.6\linewidth}
\begin{minipage}{1\linewidth}
\begin{algorithm}[H]
\caption{Stochastic Grad.\ Descent}
\label{alg:sgdrun} 
    \begin{algorithmic}
    \Require{Learning rate $\eta$}
    \Function{SGD}{$w_0, S, b, t, \err=-\infty$}
      \Let{$w$}{$w_0$}
      \For{$i \gets 1 \textrm{ to } t$}
      	\State{Sample $S'\in S,\,|S'|=b$}
          \Let{$w$}{$w - \eta \grad \EmpRisk{S'}{w} $}   %
        \If{$\BinRisk{S}{w} \leq \err $} break
        \EndIf
      \EndFor
      \State \Return{$w$}
    \EndFunction
    \end{algorithmic}
\end{algorithm}
\end{minipage}
\end{wrapfigure}
As argued above, it is reasonable to use held-out data to probe the implications of a data-dependent prior as it may give us insight into the generalization properties of $Q$.
At the same time, we may be interested in approximations to the data-dependent oracle that do not use a ghost sample.
Ordinarily, we would expect two independent runs of SGD, even on the same dataset, to produce potentially quite different weights (measured, e.g., by their $L^2$ distance) \citep{NagaKolter19c}.
\cref{fig:correlationplots} shows that, when we condition on an initial prefix of data, we dramatically decrease the variability of the learned weights. 
This experiment shows that we can predict fairly well the final weights of SGD on the full data set using only a fraction of the data set, implying that most of the variability in SGD comes in the beginning of training. 
Crucially, the two runs are \emph{coupled} in the same manner as the ghost-sample runs:
the first $\alpha$-fraction of first epoch is identical. When only a fraction of the data is available, SGD treats this data as the entire data set, starting its second epoch immediately.

\section{EMPIRICAL METHODOLOGY}
\label{sec:method}
\cref{mainexample} shows that a \doprior{} can yield tighter generalization bounds than an oracle prior.
In this section, we describe the experimental methodology we use to evaluate this phenomenon in neural networks trained by stochastic gradient descent (SGD).

\paragraph{Pseudocode.}
\cref{mainalgs} (right) describes the procedure for obtaining a PAC-Bayes risk bound on a network trained by SGD.\footnote{\cref{mainalgs} (right) uses a fixed learning rate and a vanilla SGD for simplicity, but the algorithm can be adapted to any variants of SGD with different learning rate schedules.}
Note that the steps outlined in Lines 1--3 do not change with $\sigma_P$ and therefore the best $\sigma_P$ can be chosen efficiently without rerunning the optimization. If ghost data is not used, $\gdataa$ should be replaced with $\Sa$.

To avoid choosing $\beta$, we
use a variational KL bound, described in \cref{app:varkl}, which allows us to optimize $\beta$ a posteriori for a small penalty. This PAC-Bayes bound on risk, denoted $\klLPBB{S\setminus S_{\alpha}}{Q}{P}{\delta}$, is evaluated with $\delta=0.05$ confidence level in all of our experiments during evaluation/optimization.

\begin{figure*}[t]
\centering
\includegraphics[height=32mm]{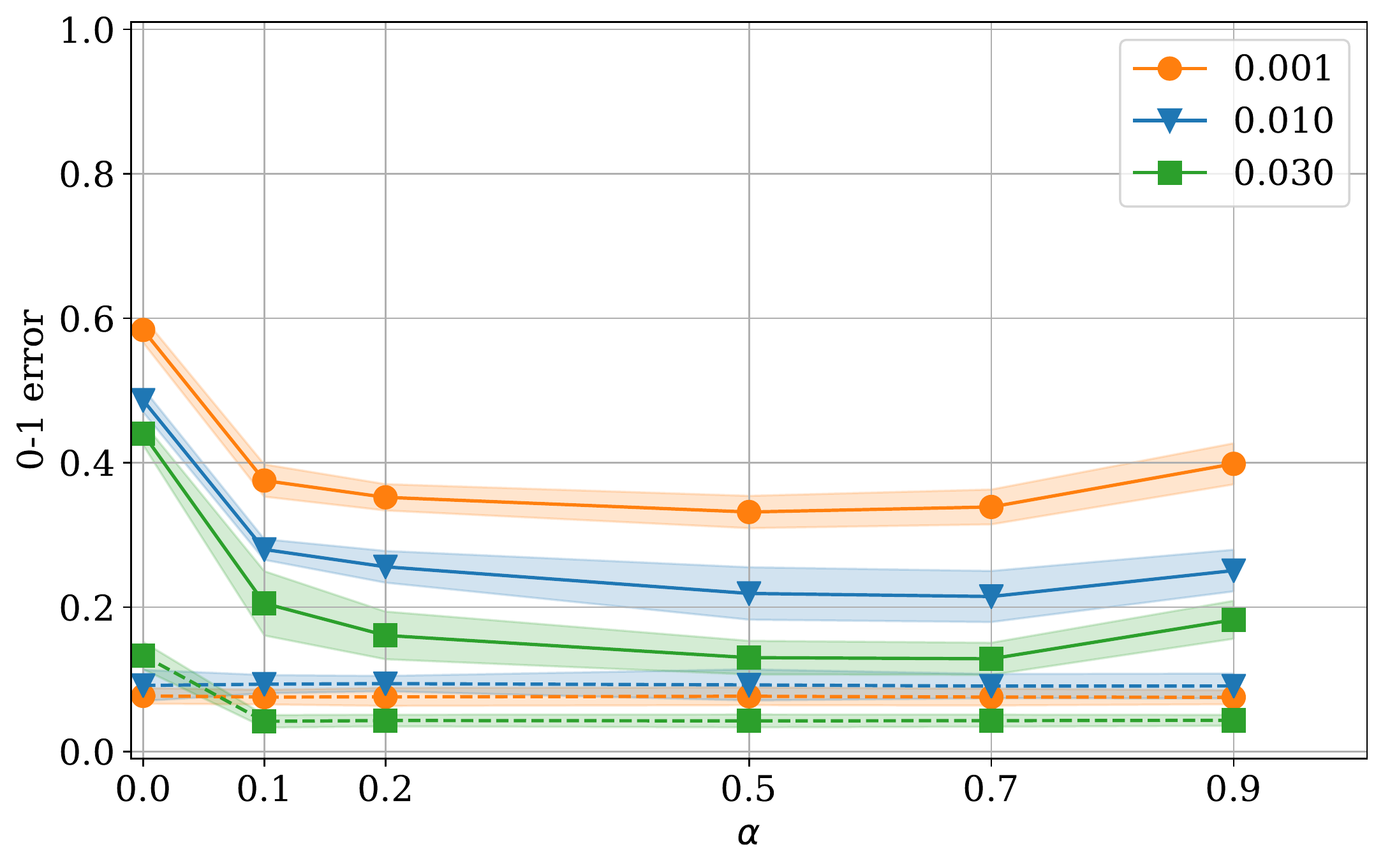}\\%
\includegraphics[height=32mm]{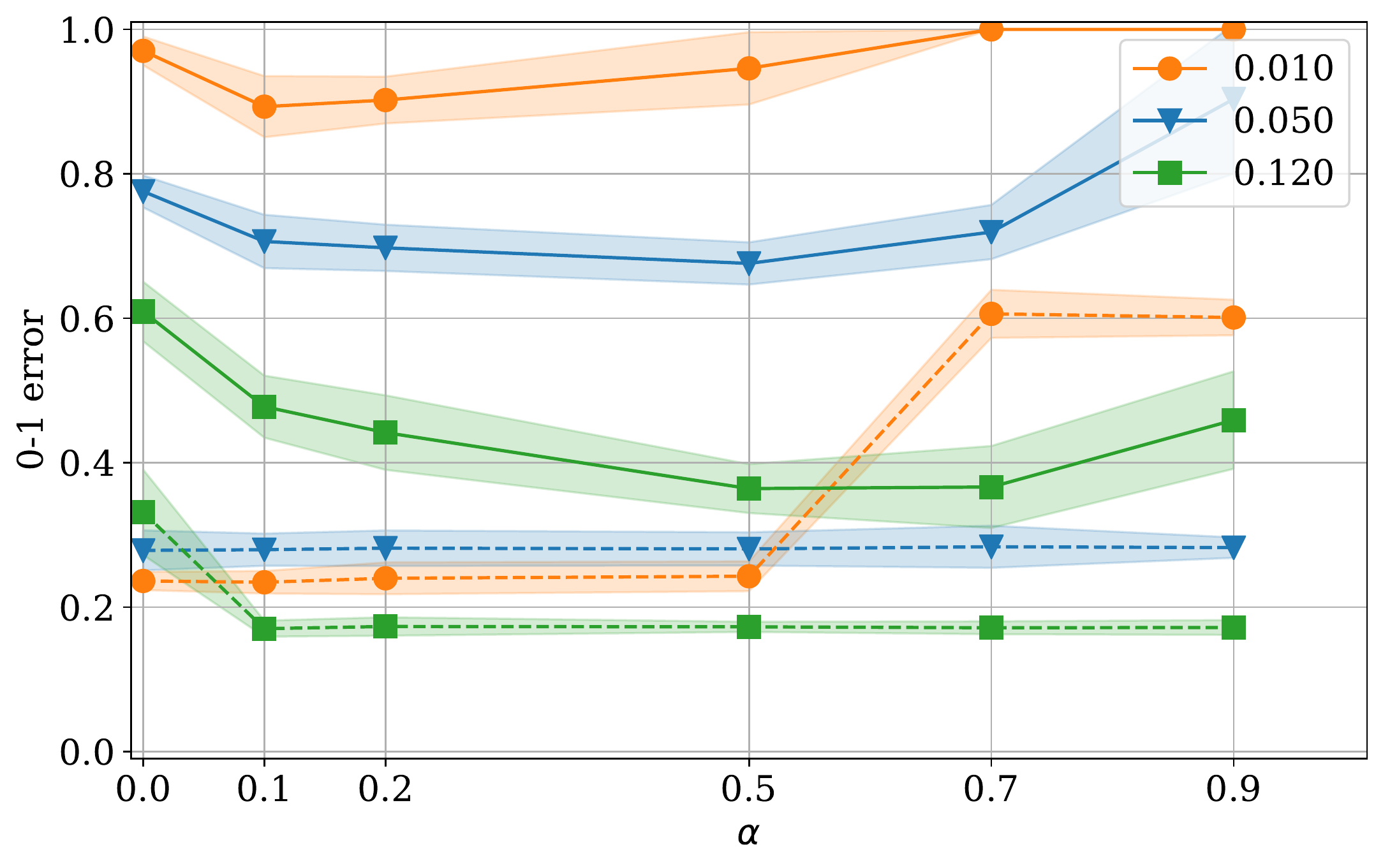}\\%
\includegraphics[height=32mm]{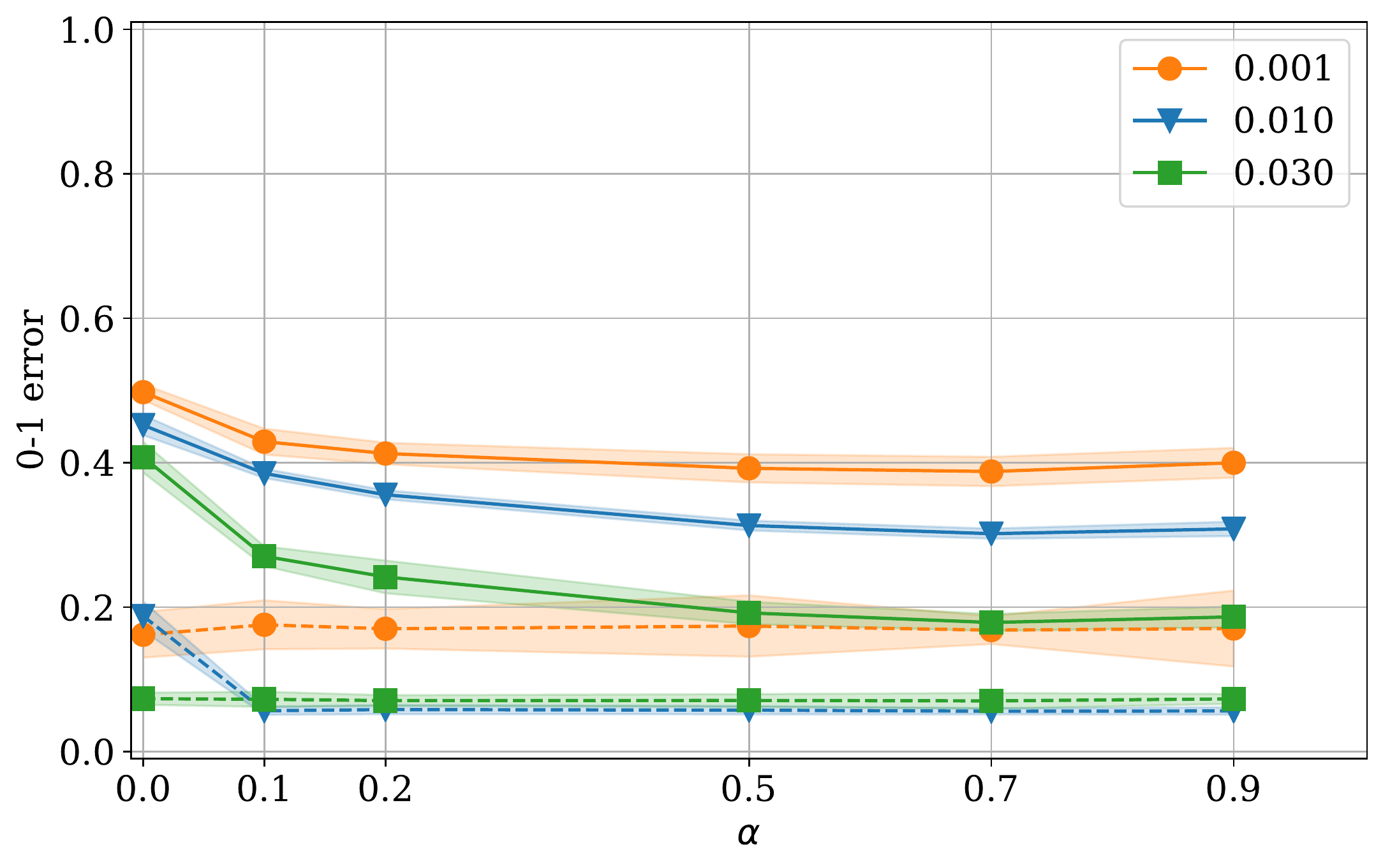}

\caption{
\textbf{top}: MNIST, LeNet-5;
\textbf{center}: Fashion-MNIST, LeNet-5;
\textbf{bottom}: MNIST, FC;
\textbf{y-axis}: error-rate;
\textbf{x-axis}: fraction $\alpha$ of the data used by the \prefix{} run of SGD to predict the weights produced by the \full{} run of SGD, $w_S$;
\textbf{dashed lines}: test error;
\textbf{solid lines}: error bound for a Gaussian Gibbs classifier $Q$, with mean $w_S$ and isotropic covariance minimizing a PAC-Bayes risk bound;
\textbf{legend}: training error used as the stopping criterion for the \full{} run of SGD.
The best error bound on MNIST ($\approx 11\%$) is significantly better than the 46\% bound by \citet{Zhou18}.
}
\label{fig:bound}
\end{figure*}

\paragraph{Datasets and Architectures.}

We use three datasets: MNIST, Fashion-MNIST and CIFAR-10. See \cref{app:datasets} for more details. 
The architectures used are described in detail in \cref{app:architectures}. For the details of the training procedure, see \cref{app:hyperparameters}.

\paragraph{Stopping criteria.}
We terminate SGD optimization in the \full{} run once the empirical error ($\BinRisk{}{}$ in \cref{mainalgs,alg:sgdrun}) measured on all of $S$ fell below some desired value $\err$, which we refer to as the stopping criteria.
We evaluate the results for different stopping criteria.

\section{EMPIRICAL STUDY OF TRAINED NETWORKS}

\paragraph{Evaluating data-dependent priors.}

A PAC-Bayes risk bound trades off empirical risk and the contribution coming from the KL term.
For isotropic Gaussian priors and posteriors, the mean component in the KL is proportional to the
 squared difference in means normalized by the effective number of training samples not seen by the prior, i.e., 
 $d(\alpha,\Sa) := \frac{\| w_S - \wwa \|^2_2 }{(1-\ks) |\datatrain| }$.
This \emph{scaled squared L2 distance} term determines the tightness of the bound when the prior variance and the posterior $Q$ and data $S$ are fixed, as the bound grows with $d(\alpha,\Sa)$.
In this section we empirically evaluate how $d(\alpha,\Sa)$ and $d(\alpha,\gdataa)$ vary with different values of $\ks$.

Our goal is to evaluate whether, on standard vision datasets and architectures, a data-dependent oracle prior can be superior to an oracle prior. Since we do not have access to an oracle prior, we approximate it by using a ghost sample $\gdataa$ with $\alpha=0$, as described in \cref{sec:coupling}. 
Data-dependent oracle priors are approximated by using a combination of training samples and ghost samples.

Our experimental results on MNIST and Fashion-MNIST appear in \cref{fig:l2distanceplots}, where we plot $d(\alpha,\Sa)$ and $d(\alpha,\gdataa)$.
The results suggest that the value of $\alpha$ minimizing $d(\alpha,\gdataa)$ is data- and architecture-dependent.
The optimal prefix size for MNIST, FC minimizing $d(\alpha,\Sa)$ is $\alpha>0.2$.
For MNIST, LeNet-5 and Fashion-MNIST, LeNet-5, the optimal $\alpha$ is between $0$ and $0.1$.
We found that batch size affects the optimal $\alpha$, whether on \prefix{} or ghost data. As one might expect, the best $\alpha$ is larger for smaller batch sizes. We hypothesize that this is due to increased stochasticity of SGD.

Interestingly,
at larger values of $\alpha$ we observe that the gap between $d(\alpha,\Sa)$ and $d(\alpha,\gdataa)$ closes.
This happens in all three experimental setups by $\alpha = 0.4$: we observe that the prior mean obtained with $\Sa$ training data alone is as close to final SGD weights as the prior mean obtained with $\gdataa$.

\paragraph{Generalization bounds for SGD-trained networks.}

We apply data-dependent priors to obtain tighter PAC-Bayes risk bounds for SGD-trained networks.
We do not use ghost data in these experiments, as oracle priors are inaccessible in practice.
Thus the prior mean is obtained by the \prefix{} run on prefix data alone.
See \cref{mainalgs} (right) and \cref{sec:method} for the details of the experiment.

From the data in \cref{fig:bound}, it is apparent that $\alpha$ has a significant impact on the size of the bound. In all of the three networks tested, the best results are achieved for $\alpha>0$.

One of the clearest relationships to emerge from the data is the dependence of the bound on the stopping criterion: The smaller the error at which the \full{} run was terminated, the looser the bound. This suggests that the extra optimization introduces variability into the weights that we are not able to predict well.
In \cref{sec:oraclevar}, we use oracle bounds to quantify limits on how much tighter these generalization bounds could be, were we able to 
optimize a diagonal prior variance. The results suggest that a diagonal prior offers little advantage over an isotropic prior.

\begin{figure}[t]
\includegraphics[width=.7\linewidth]{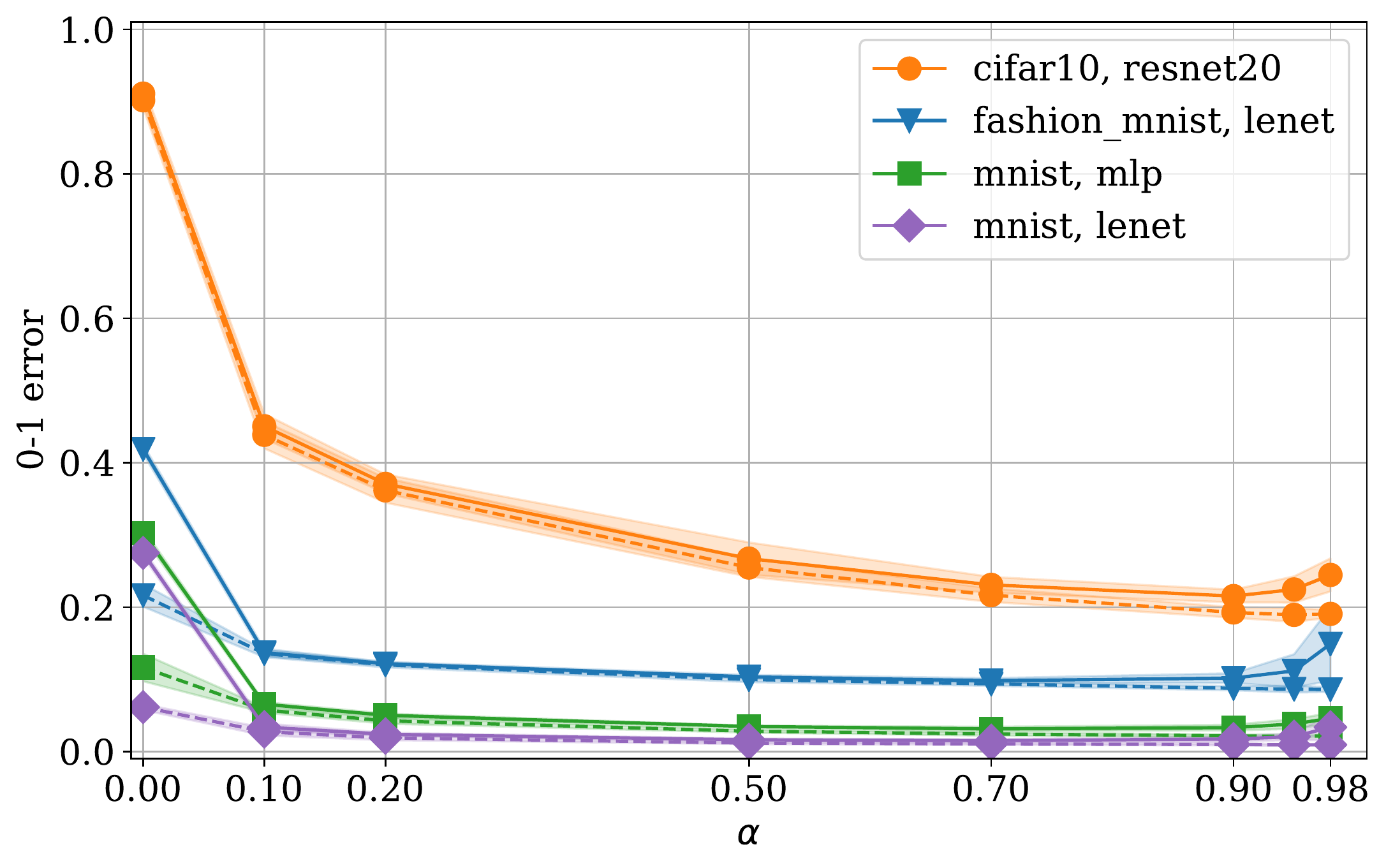}
\caption{
\textbf{Y-axis}: error-rate;
\textbf{x-axis}: fraction $\alpha$ of the data used to learn the prior mean;
\textbf{dashed lines}: test error;
\textbf{solid lines}: bound on the error of a Gaussian Gibbs classifier whose mean and diagonal covariance are learned by optimizing the bound surrogate;
\textbf{legend}: dataset and network architecture.
For each scenario, under the optimal $\alpha$, the bound is tight and test error is within a few percent of standard SGD-trained networks.
}
\label{fig:directboundopt}
\end{figure}

\paragraph{Direct risk bound minimization.}
\label{sec:direct}

One of the dominant approaches to training Gaussian neural networks is to minimize the evidence lower bound (ELBO), 
which essentially takes the same form as the bound in \cref{pacbayeslinear}, but with a different relative weight on the KL term.
Here, we optimize a PAC-Bayes bound using our data-dependent prior methodology which can be related to empirical Bayes approaches.
The details of the algorithm are outlined in \cref{mainalgs}, left, where $\diffLPBB{S\setminus S_{\alpha}}{Q}{P}{\delta}$ denotes a PAC-Bayes bound computed with differentiable surrogate loss. We perform experiments on 3 different datasets and architectures (see \cref{app:drbm} for further details).

\cref{fig:directboundopt} presents the error of the posterior $Q$ (dashed line) optimized using \cref{mainalgs} with different values of $\alpha$. 
It is apparent from the figure that for all the networks and datasets tested, 
the error of $Q$ drops dramatically as $\alpha$ increases, all the way up to around $\alpha=0.9$.
Note that $Q$ with the optimal $\alpha$ achieves very high performance even compared to state-of-the-art networks and at the same time comes with a valid guarantee on error. 
For example, ResNet20 (without data augmentation and weight decay) trained on CIFAR10 achieved error of around $0.16$, and the best-performing $Q$ in \cref{fig:directboundopt} gets an average error of $\approx 0.2$ with a bound $\approx 0.23$ that holds with 0.95 probability.

\printbibliography

\clearpage

\appendix

\onecolumn

\section*{Supplementary Materials}

\startcontents[sections]
\printcontents[sections]{l}{1}{\setcounter{tocdepth}{1}}

\newcommand{\Sp}{S_{J}}
\newcommand{\Sr}{S_{\bar J}}

\section{Proof of \cref{mainthm}}
\label{app:mainthm}

The proof is straightforward, following essentially from definitions. We give it here for completeness.

We pause to make two remarks about $J$:
\begin{enumerate}

\item We do not require $J$ to have any particular distribution and so, e.g., $J$ could be uniformly distributed among subsets of cardinality $\alpha n$ or could be a.s.\ nonrandom and equal to $[m]$. 
\item The statement that $\Pr[\hat h|S,J] = Q(S)$ a.s.\ implies that $\Pr[\hat h|S] = Q(S)$ a.s.\ and that $\hat h$ is independent of $J$, both marginally and conditionally on $S$. 
Informally,
any randomness in $J$ plays no role in the determination of $\hat h$.
\end{enumerate}
Let $\bar J = [n] \setminus J$,
fix $\PF$,
and consider the linear PAC-Bayes bound \emph{based on} $\EmpRisk{\Sr}{Q(S)}$, i.e., where we
use the data in $\Sr$ to estimate the risk of $Q(S)$.
By the linear PAC-Bayes theorem, we are permitted to choose our prior based on $S_J$, since $S_J$ is independent of $\Sr$.
In fact, we can also choose our prior knowing  $J$, due to the independence outlined above in the second remark.

Conditionally on $J$, the expected value of
the linear PAC-Bayes bound under the data-dependent oracle prior
is the infimum
\[
&\inf_{P \in Z^{m} \to \PF} 
      \EE_{S}[ \LPBB{\beta}{\Sr}{Q(S)}{P(S_J)}{\delta}]
\\&= \EE_{S}  [ \beta^{-1} \EmpRisk{\Sr}{Q(S)}] + 
\inf_{P \in Z^{m} \to \ProbMeasures{\HS}} 
  \frac { \EE_S[\KL{Q(S)}{P(\Sp)} ]+ \log \frac {1}{\delta} }{2 \beta(1-\beta) (1-\alpha) n}
\\&=
\EE_S[ \beta^{-1} \EmpRisk{\Sr}{Q(S)}]
	+ \frac { \EE_S[\KL{Q(S)}{\EE[Q(S)|\Sp]} ] + \log \frac {1}{\delta} }{2 \beta(1-\beta) (1-\alpha)n}
\\&= 
\EE_S[ \beta^{-1} \EmpRisk{\Sr}{Q(S)}]
	+ \frac { \dcMI[\PF]{J}{\hat{h}}{S}{\Sp} + \log \frac {1}{\delta} }{2 \beta(1-\beta) (1-\alpha) n}.
\]
Here $\EE_{S}$ denotes expectation over $S \sim \Dist^n$, conditional on $J$.
Note that the optimal prior here depends on $J$. We arrive at the unconditional expected value of the bound by taking expectations also over $J$, which changes the \emph{disintegrated} (conditional) mutual information $\dcMI[\PF]{J}{\hat{h}}{S}{\Sp}$ into a (conditional) mutual information $\cMI[\PF]{\hat{h}}{S}{\Sp,J}$, that is no longer a random quantity.

It follows immediately that the data-dependent risk bound is tighter, in expectation, that the bound based on $J = \emptyset$, when
\[\begin{split}
&(1-\beta)\, \EE[ \EmpRisk{S}{Q(S)} ]
	+ \frac { \MI[\PF]{\hat{h}}{S} + \log \frac {1}{\delta} }{2 n}
\\&\qquad	> 
(1-\beta)\, \EE[ \EmpRisk{\Sr}{Q(S)}]
	+ \frac { \cMI[\PF]{\hat{h}}{S}{\Sp,J} + \log \frac {1}{\delta} }{2 (1-\alpha) n}.
\end{split}
\]
The statement of the proposition is obtained by simple manipulations. The above inequality is equivalent to 
\[
\begin{split}
&\frac { \MI[\PF]{\hat{h}}{S} + \log \frac {1}{\delta} }{2 n} - \frac { \cMI[\PF]{\hat{h}}{S}{S_J,J} + \log \frac {1}{\delta} }{2 (1-\alpha) n}
\\&\qquad > 
(1-\beta)\, \EE[ \EmpRisk{\Sr}{Q(S)} - \EmpRisk{S}{Q(S)}].
\end{split}
\]
Rewriting the left-hand side,
\[
\begin{split}
&\frac { \MI[\PF]{\hat{h}}{S} + \log \frac {1}{\delta} }{2 n} - \frac { \cMI[\PF]{\hat{h}}{S}{S_J,J} + \log \frac {1}{\delta} }{2 (1-\alpha) n}
\\&\qquad= 
\frac 1 2 \left (
\frac { \MI[\PF]{\hat{h}}{S}}{n} - \frac { \cMI[\PF]{\hat{h}}{S}{S_J,J} }{(1-\alpha) n} 
\right)
- \frac{ \log \frac {1}{\delta} }{2 n} \left ( \frac {\alpha }{1-\alpha } \right ).
\end{split}
\]
Therefore, we prefer a data-dependent prior based on $J$ when
\[
\begin{split}
&\left (
\frac { \MI[\PF]{\hat{h}}{S}}{n} - \frac { \cMI[\PF]{\hat{h}}{S}{S_J,J} }{(1-\alpha) n} 
\right)
\\&\qquad > 
2(1-\beta)\, \EE[ \EmpRisk{\Sr}{Q(S)} - \EmpRisk{S}{Q(S)}]
+ \frac{ \log \frac {1}{\delta} }{n} \left ( \frac {\alpha }{1-\alpha } \right ).
\end{split}
\]
The result follows by the definition of the information rate gain and excess bias. Note that, if $J$ is (a.s.) nonrandom, then 
$\cMI[\PF]{\hat{h}}{S}{S_J,J} $ is simply $\cMI[\PF]{\hat{h}}{S}{S_J} $.

\section{Proof of \cref{keyclaim}}
\label{app:fullproof}

We begin with a proof sketch.

\begin{proof}[Sketch]
With $J$ and $\beta$ fixed,
the minimization over $P(S_J)$ meets the hypotheses of \cref{oracleprior} and so we may simplify the objective
by taking $P(S_J) = \EE[Q(S)|S_J] = \Pr[W|S_J]$. The KL term then becomes a conditional mutual information $\cMI{W}{S \setminus S_J}{S_J}$.
Due to linearity of expectation, we may then optimize $\beta$ explicitly,
leaving only a minimization over subsets $J$,
\begin{equation*}
\inf_{J \subseteq [n]} \,
 \bigg( \Phi(J) := R(J) + C(J) + \sqrt{2 R(J) C(J) + C^2(J) } \bigg)
\end{equation*}
where
$R(J) = \EE [ \EmpRisk{S \setminus S_J }{Q} ]$
and
$C(J) = ({ \cMI{W}{S \setminus S_J}{S_J} + \log \frac {1}{\delta} } ) / {|S \setminus S_J| }$.

One can show that $\cMI{W}{S \setminus S_J}{S_J} = \frac{D}{2} \ln \phi_{\bar J} / \kappa$,
where $\phi_{\bar J}$ is the variance contribution from $S \setminus S_J$ and $\xi$.
Using sub-Gaussian and sub-exponential tail bounds, one can establish that
$R(J) \le \oR = \exp\{{-D/16}\} + \exp \{- \tau^2/(4\phi_{[n]} \,\sigma^2)\}$,
where $\phi_{[n]}$ is due to variance in $S$, $\xi$, and $\tau = (\sum_{i=1}^{n} \eta_i) \norm{u}^2$.

Choosing $n=100$, $D=1000$, $\sigma=8$, $\kappa=4$, $\tau = 64$, and $\delta=0.05$,
we obtain $\Phi(\emptyset) \ge 2C(\emptyset) \approx 1.1$,
while $\min_{J} \Phi(J) \lessapprox 0.15$.
Our upper bound is achieved by $J=[24]$, i.e., by using the initial 24 data points to obtain a data-dependent (oracle) prior.
\end{proof}

\subsection{Complete proof and bounds on the objective}
We now provide a complete rigorous proof.
For subsets $J$ (of $[n]$),
let $\bar J = [n] \setminus J$;
let $\etas{J}{p} = \sum_{i \in J} \eta_i^p$ for $p \in \{{1,2}\}$;
let $\phi_{J} =  \etas{J}{2} \sigma^2 / D  + \kappa$;
and let $\phi_{-i} = \phi_{[n]\setminus \{{i}\}}$.

By \cref{oracleprior} and linearity of expectation,
for every subset $J$ and $\beta \in (0,1)$,
\cref{oracleprior} implies that
the optimal prior is $P_J(\Sp) = \Pr[W|\Sp]$, and so we can simplify \cref{theinf} by choosing this prior.
In particular, now $\EE [\KL{Q}{P_J(\Sp)}] = \cMI{W}{\Sr}{\Sp}$.

Define
$R(J) = \EE [ \EmpRisk{\Sr}{Q} ]$
and
$C(J) = ({ \cMI{W}{\Sr}{\Sp} + \log \frac {1}{\delta} } ) / {|\Sr| }$.
By linearity of expectation,
we can remove the infimum over $\beta \in (0,1)$ by explicit minimization.
As a result, we see that \cref{theinf} is equivalent to
\[\label{optbound}
\inf_{J \subseteq [n]} \,
 \underbrace{R(J) + C(J) + \sqrt{2 R(J) C(J) + C^2(J) }}_{\Phi(J)}.
\]

Pick some $J \subseteq [n]$.
Then the optimal prior conditioned on $\Sp$ is
\[
P_J(S_J) = \EE[Q(S)|\Sp] = \delta_{\etas{[n]}{1}\, \vu} \otimes N_J,
\]
where $N_J = \Normal {\sum_{i\in J} \eta_{i} y_{i} \vx_{i,2}}{ \phi_{\bar J}\, I_D}$.
Let $\psi(r) = r - 1 - \ln r$ for $r > 0$.
Then
\[
\KL{Q(S)}{P_J(\Sp)}
= D \psi( \kappa / \phi_{\bar J}) / 2
+
\frac 1 {2 \phi_{\bar J}}
\sum_{j=1}^{D}
 \big (\sum_{i \not\in J} \eta_{i} y_{i} \vx_{i,2,j} \big)^2 .
\]
Taking expectations, %
\[
\cMI{W}{\Sr}{\Sp}
= \EE [\KL{Q(S)}{P_J(\Sp)}]
&= \frac{D}{2} \Big ( \psi( \kappa / \phi_{\bar J})
+ \frac { \sigma^2 \, \etas{\bar J}{2} / D} { \phi_{\bar J} }  \Big )
\\&= \frac{D}{2} \Big ( \psi( \kappa / \phi_{\bar J}) + (1-\kappa/\phi_{\bar J}) \Big )
\\&= \frac{D}{2} \ln \phi_{\bar J} / \kappa.
\]

It remains to control the empirical risk term.
To that end, pick $i \in [n]$
and let $\tau = \etas{[n]}{1} \norm{u}^2$.
Then
\[
\EE \loss(W,z_i)
= \Pr [ y_i \ip{W}{\vx_i} \le 0] 
&= \Pr [ \tau + y_i \ip{w_{n,2}}{\vx_{i,2}} + y_i \ip{\xi}{\vx_{i,2}} \le 0],
\]
where
\[
y_i \ip{w_{n,2}}{\vx_{i,2}} = \eta_i \norm{\vx_{i,2}}^2 + \sum_{j\neq i} \eta_j y_i y_j \ip{\vx_{j,2}}{\vx_{i,2}}.
\]
Rearranging and exploiting the chain rule of conditional expectation and symmetry of the normal distribution,
\begin{align*}
\EE \loss(W,z_i)
= \EE \Pr^{\vx_{i,2}} [ \sum_{j\neq i} \ip{\eta_j\,\vx_{j,2}}{\vx_{i,2}} + \ip{\xi}{\vx_{i,2}} \ge \tau + \eta_i \norm{\vx_{i,2}}^2]
,
\end{align*}
where the conditional probability is a tail bound on a univariate Gaussian with mean zero
and variance $\norm{\vx_{i,2}}^2 \phi_{-i}$.

Applying the standard (sub-)Gaussian tail bound,
\[
\EE \loss(W,z_i)
\le \EE \exp \bigg \{{- \frac 1 2 \frac{ (\tau + \eta_i \norm{\vx_{i,2}}^2)^2 }{\norm{\vx_{i,2}}^2 \phi_{-i}} } \bigg \}
\le \EE \exp \bigg \{{
          - \frac{\tau^2}{{2\norm{\vx_{i,2}}^2 \phi_{-i}}}
           } \bigg \},
\]
where the last inequality is crude, but suffices for our application.
Note that  $D \norm{\vx_{i,2}}^2/\sigma^2$ is a chi-squared random variable with $D$ degrees of freedom,
hence sub-exponential.
Indeed, with probability at least $1-c$,
\[
D \norm{\vx_{i,2}}^2/\sigma^2 \le D + 2\sqrt{D \log (1/c)} + 2\log(1/c).
\]
Rearranging,
\[
\norm{\vx_{i,2}}^2
&\le \frac{\sigma^2}{D}(D + 2\sqrt{D \log (1/c)} + 2\log(1/c))
\\&\le \sigma^2 (1 + 4\sqrt{\log (1/c)/D}) =: B(c),
\]
where the second inequality holds assuming $c \ge \exp \{{-D}\}$,
which we will ensure from this point on.
So
\[
\EE \loss(W,z_i) \le \inf_{c \ge e^{-D}} \bigg \{
\frac 1 c +
\Big (1 - \frac 1 c \Big) \exp \{ - \tau^2/(2 \phi_{-i} B(c)) \}.
\bigg \}.
\]
Taking $c = \exp\{{-D/16}\}$,
we have $B(c) = 2\sigma^2$.
Then, using $\phi_{-i} \le \phi_{[n]}$,
\[
\EE \loss(W,z_i)
\le \underbrace{\exp\{{-D/16}\} + \exp \{- \tau^2/(4\phi_{[n]} \,\sigma^2)\}}_{\oR}.
\]
We may now obtain a bound
\[
R(J) = \EE \EmpRisk{\Sr}{Q(S)}
= \frac 1 {n-|J|} \sum_{i \not\in J} \EE \loss(W,z_i)
\le \max_{i \not \in J} \, \EE \loss(W,z_i) = \oR.
\]

Thus
\[
\Phi(J) \le \overline{R} + C(J) + \sqrt{2 \overline{R}\,C(J) + C(J)^2}
\]
At the same time,
we have $\Phi(J) \ge 2 C(J)$ for all $J \subseteq [n]$. (\textbf{Note that these two bounds are used to produce \cref{fig:theorybound}.})

In particular, noting $\log 1/\delta > 0$,
\[
\Phi(\emptyset) \ge \frac{D}{m} \ln \frac {\sigma^2 \etas{[n]}{2} / D  + \kappa }{\kappa}.
\]
The result can be seen to follow from these bounds by evaluation using the particular values.
In particular, one can see that taking $J$ to be a nonempty initial segment of $[n]$, we have $\Phi(J) < 2C(\emptyset) \le \Phi(\emptyset)$.

\section{Analytic form of the KL for an approximate data-dependent oracle bound}
\label{app:approx}

In this section, we explore one possible analytic bound for a KL term for a PAC-Bayes bound,
based on the setup in \cref{sec:sgdddpriors}. 
We assume $\trace(\covf)$ and $\det(\covf)$ are nonrandom. 
In an application, one would have to cover a set of possible values to handle the  random case.

The KL divergence between Gaussians
$Q(\RS,S) = \Normal{\wwf}{\covf}$
and $P= \Normal{\wwa}{\cova}$
 takes the form
\[\label{klgaussians}
2\KL{Q(\RS,S)}{\ddprior}
=
\underbrace{
\norm{\wwf - \wwa}_{\cova^{-1}}^2 \vphantom{\ln \frac {\det \cova}{\det \covf}}}_{\textrm{mean component}}
+
\underbrace{
\trace(\cova^{-1} (\covf)) - \pdim + \ln \frac {\det \cova}{\det \covf}.
}_{\textrm{variance component}}
\]
Specializing to an isotropic prior, i.e.,
$\cova = \pvar I$,
we obtain
\[
2\KL{Q(\RS,S)}{\ddprior}
=
\underbrace{
\frac 1 {\pvar }\norm{\wwf - \wwa}^2 \vphantom{\ln \frac {\det \cova}{\det \covf}}}_{\textrm{mean component}}
+
\underbrace{
\frac 1 {\pvar }\trace(\covf) - \pdim + p \ln \pvar - \ln \det \covf .
}_{\textrm{variance component}}
\]

\newcommand{\ccov}[2]{\mathrm{cov}^{#1}(#2)}
Note that
\[
\trace(\ccov{\Sa,U}{\wwf}) = \inf_{\wwa} \ \cEE{\Sa,U}{\norm{\wwf - \wwa}^2 }.
\]
Consider
\[
\pvar = \frac 1 p \big (  \trace(\ccov{\Sa,U}{\wwf}) + \trace(\covf) \big ).
\]
Substituting above,
for some random variable $Z$ such that $\cEE{\Sa,U}{Z}=1$,
\[
2\KL{Q(\RS,S)}{\ddprior}
&= Z p - p + p \ln \Big \{ \frac {\frac 1 p  \trace(\ccov{\Sa,U}{\wwf}) + \frac 1 p \trace(\covf) } {(\det \covf)^{1/p}}  \Big \}
\\&\le Z p - p + p\frac {\frac 1 p\trace(\ccov{\Sa,U}{\wwf}) +  \frac 1 p \trace(\covf) - (\det \covf)^{1/p} } {(\det \covf)^{1/p}} .
\]

Taking expectations, conditional on $\Sa,U$,
\[
2 \cEE{\Sa,U}{\KL{Q(\RS,S)}{\ddprior}}
\le \frac {\trace(\ccov{\Sa,U}{\wwf})} {(\det \covf)^{1/p}}
        + p \frac { \frac 1 p \trace(\covf) - (\det \covf)^{1/p} } {(\det \covf)^{1/p}} .
\]

Further, if we assume $\Sigma = \sigma I$, then
\[
\cEE{\Sa,U}{\KL{Q(\RS,S)}{\ddprior}}
\le \frac 1 {2 \sigma} \trace(\ccov{\Sa,U}{\wwf}).
\]

\section{Variational KL bound}
\label{app:varkl}

The linear PAC-Bayes bound requires one to specify a value of $\beta$. 
For a particular posterior kernel $Q$, the optimal value of $\beta$ depends on the likely value of the empirical risk term. However, the value of $\beta$ must be chosen independently of the data used to evaluate the bound.

In the proof of \cref{keyclaim} in \cref{app:fullproof},
the linear PAC-Bayes bound is optimized, in expectation. Since the expected value of the bound is independent of the data, and since the constant $\beta$ can be pulled outside the expectations, we can choose the value of beta that minimizes the bound in expectation. The result is \cref{optbound}, with $C(J)$ defined in terms of an expected KL, as the mutual information appears only when the prior is chosen to be the oracle prior.

In this section, we describe how the bound due to \citet{Maurer04} can be approximated to reveal a high-probability tail bound with the same form as if we optimized $\beta$. The cost is a $O(\log \sqrt{m}/m)$ term.

Let $\Bernoulli{p}$ denote the Bernoulli distribution on $\{0,1\}$ with mean $p$.
For $p,q \in [0,1]$, we abuse notation and define
\begin{equation*}
\KLbin{q}{p} \defas
\smash{ \KL{\Bernoulli{q}}{\Bernoulli{p}} = q \ln \frac q p + (1-q) \ln \frac {1-q}{1-p}. }
\end{equation*}

The following PAC-Bayes bound for bounded loss is due to \citet{Maurer04}.
The same result for 0--1 loss was first established by \citet{LS01}, building
off the seminal work of \citet{PACBayes}.
See also \citep{LangfordPHD} and \citep{Catoni}.

\begin{theorem}[PAC-Bayes; {\citealt[Thm.~5]{Maurer04}}]
\label{pacbayes}
Under bounded loss $\loss \in [0,1]$,
for every $\delta > 0$, $m \in \Nats$, distribution $\Dist$ on $Z$, and distribution $P$ on $\HS$,
\begin{equation}
\PPr{S \sim \Dist^m}
    \Bigl (
      (\forall Q)\
      \KLbin{\EmpRisk{S}{Q}}{\Risk{\Dist}{Q}} \le \frac {\KL{Q}{P} + \ln \frac{2 \sqrt{m}}{\delta}}{m}
    \Bigr ) \ge 1-\delta.
\end{equation}
\end{theorem}

One can recover the bound by \citet{PACBayes} via Pinsker's inequality,
resulting in a (looser) bound on $\abs{ \EmpRisk{S}{Q} - \Risk{\Dist}{Q}}$.
Maurer's bound behaves like the bound in \cref{pacbayeslinear},
except that it holds for all $\beta$ simultaneously, at the cost of a $\frac 1 m \log \sqrt{m}$ term.

\subsection{Inverting the KL bound}
\label{app:inverting}

Here we derive a novel PAC-Bayes bound that is an upper bound on the inverted KL bound (\cref{pacbayes}) and that is used during optimization in our empirical work.
The bound is the piecewise combination of two bounds. In independent work, \citet{rivasplata2019pac} derive the first of the two parts, which they call a ``quad bound''. The second part is a consequence of Pinsker's inequality.

\begin{theorem}[Variational KL bound]\label{variational_kl_bound}
With probability at least $1-\delta$ over $S$,
\[
\label{varklbound}
\Risk{\Dist}{Q} \leq \min
\begin{cases}
\EmpRisk{S}{Q} + B + \sqrt{ B(B+2 \EmpRisk{S}{Q}) }   ,\\
\EmpRisk{S}{Q} + \sqrt{ \frac B 2 }  ,
\end{cases}
\]
where
\[
B = \frac{ \KL{Q}{P} + \log \frac{2\sqrt{m}}{\delta} }{|S|}.
\]
\end{theorem}
The variational KL bound takes the minimum value of the \emph{\momentbound{}} (top) and the \emph{\Pinskerbound{}} (bottom).
\begin{proof}
Let $\KLbin{\EmpRisk{S}{Q} }{\Risk{\Dist}{Q}}$ be KL between two Bernoulli random variables with success probabilities $\EmpRisk{S}{Q}$ and $\Risk{\Dist}{Q}$, respectively.
Then by \cref{pacbayes}, with probability greater than $1-\delta$,
\[\label{eq:seegerbound}
\KLbin{\EmpRisk{S}{Q} }{\Risk{\Dist}{Q}} \leq \frac{\KL{Q}{P} +  \log \frac{2 \sqrt{|S|}}{\delta} }{|S|}.
\]
Let $B$ denote the right hand side of the inequality.
By Donsker--Varadhan we get
\[
\KLbin{\EmpRisk{S}{Q} }{\Risk{\Dist}{Q}} \geq \lambda \EmpRisk{S}{Q} - \log \EEE{x \sim \text{Ber}(\Risk{\Dist}{Q} ) } [e^{\lambda x}]
\]
for any $\lambda$. The final term is the moment generating function of a Bernoulli random variable and so
\[
\KLbin{\EmpRisk{S}{Q} }{\Risk{\Dist}{Q}} \geq \lambda \EmpRisk{S}{Q}  - \log (1- \EmpRisk{S}{Q}  + \EmpRisk{S}{Q}  e^{\lambda}).
\]
We can use this lower bound on $\KLbin{\EmpRisk{S}{Q} }{\Risk{\Dist}{Q}} $ in \cref{eq:seegerbound}. After rearranging, we obtain
\[
- \Risk{\Dist}{Q} (1-e^{\lambda} ) \geq e^{\lambda \EmpRisk{S}{Q}  - B} -1.
\]
Take $\lambda\leq 0$. Then
\[\label{eq:riskboundwithlambda}
\Risk{\Dist}{Q} \leq \frac{1 - e^{\lambda \EmpRisk{S}{Q}  - B} }{ 1 - e^{\lambda}  } .
\]
Using the inequality $1-e^{-x} \leq -x$ in the numerator of \cref{eq:riskboundwithlambda}, we finally arrive at
\[\label{eq:topbound}
\Risk{\Dist}{Q} \leq \EmpRisk{S}{Q}  + B + \sqrt{B(B+2 \EmpRisk{S}{Q} )}.
\]
Also, note that by Pinsker's inequality,
\[
\KLbin{\EmpRisk{S}{Q} }{\Risk{\Dist}{Q}}  \geq 2 (\EmpRisk{S}{Q} - \Risk{\Dist}{Q})^2,
\]
and so
\[\label{PinskerBound}
\Risk{\Dist}{Q} \leq \EmpRisk{S}{Q} + \sqrt{\frac B 2}.
\]
Both \cref{PinskerBound} and \cref{eq:topbound} are upper bounds on risk obtained from the inverted kl bound. Taking the minimum of the two bounds gives us the final result.
\end{proof}

\begin{figure}[t!]
\centering
\includegraphics[width=.7\linewidth]{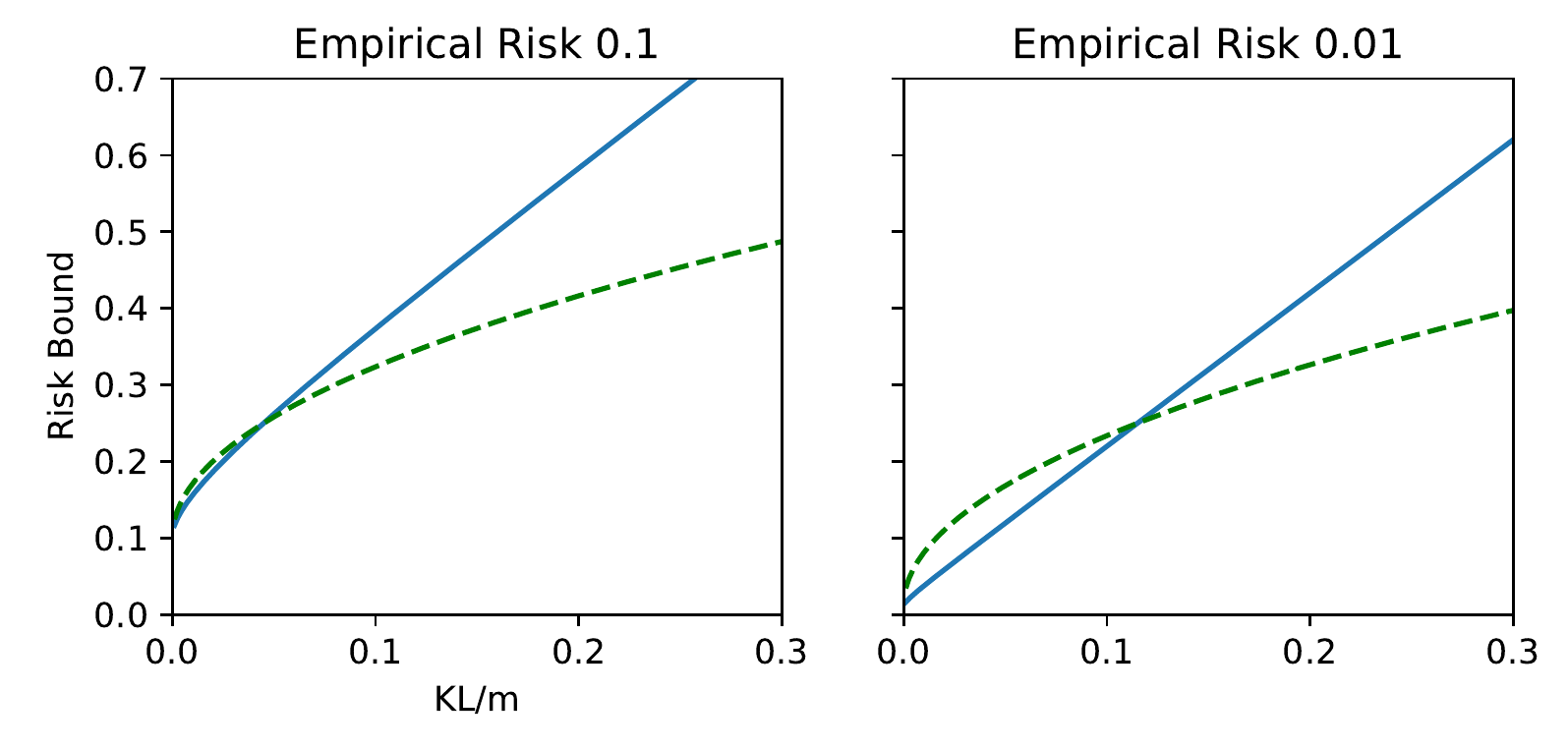}
\caption{
The two components of Variational KL bound visualized for $\EmpRisk{S}{Q} = 0.1$ (left) and $\EmpRisk{S}{Q} = 0.01$ (right).
\textbf{Blue solid line:} \momentbound{};
\textbf{Green dashed line:} \Pinskerbound{}.
The inverted KL bound on the risk is the minimum of the two lines.
}
\label{fig:invertedklbound}
\end{figure}

The inverted KL bound is visualized in \cref{fig:invertedklbound}. We see that depending on the empirical risk and KL, either the moment or the \Pinskerbound{} is tighter.
The inverted KL bound is the minimum of the two and so is tight in both regimes.
By taking the minimum of two bounds, we obtain a bound this is tighter over a wider range of values for the empirical risk and KL terms. 

\raggedbottom

\section{Experimental details: datasets}
\label{app:datasets}

We use three datasets in our experiments:
1) The MNIST dataset~\citep{MNIST}, which consists of $28\times28$ grayscale images of handwritten decimal digits. 2) The Fashion-MNIST dataset~\citep{fashionmnist}, which consists of $28\times28$ grayscale images each associated with one of 10 categories (clothing and accessories). 3) The CIFAR-10 dataset~\citep{cifar10}, which consists of $32\times32$ RGB images each associated with one of ten categories (airplane, automobile, bird, etc.). For all datasets we use the standard training and test splits. This results in 60,000 training data for MNIST and Fashion-MNIST, 50,000 training data for CIFAR-10, and 10,000 test data for all three datasets. For CIFAR-10 we standardize all images according to the training split's statistics.

\section{Experimental details: architectures}
\label{app:architectures}

We use fully connected feed-forward multilayer perceptrons with ReLU activations for MNIST. We study networks with architecture 784--600--600--10 (featuring two hidden layers) in order to compare to \citet{rivasplata2019pac}. Such a network has 837,610 parameters.

We also borrow the modified LeNet-5 architecture used by \citet{Zhou18} in order to compare our bounds on SGD-trained classifiers. The network has 431,080 parameters. We use this architecture for MNIST and Fashion-MNIST.

We use the ResNet-20 architecture~\citep{he2016deep} for CIFAR-10. It has 269,722 parameters. For consistency with the other experiments, we use neither data augmentation nor weight decay.

\section{Experimental details: training details}
\label{app:hyperparameters}

The bounds are evaluated on the 0--1 loss, which is not differentiable. To enable gradient-based optimization, we replace this with the cross entropy loss divided by the $\log$ number of classes, which gives a tight upper bound on the 0--1 loss.

We use SGD with momentum as the optimizer. We use one learning rate for the \prefix{} and \full{} runs and another, lower learning rate for the bound optimization. For experiments on MNIST and Fashion-MNIST, the momentum is 0.95 and the batch size is 256. For MNIST, the learning rate for the \prefix{} and \full{} runs is 0.003 and the learning rate for bound optimization is 0.0003; for Fashion-MNIST, they are respectively 0.01 and 0.003. We sweep over the prior variance $\sigma_P \in \{\SI{3e-8}, \SI{1e-7}, \SI{3e-7}, \dots, \SI{1e-2}\}$.
Via a union bound argument, our hyperparameter sweeps contribute a negligible amount to the bounds.

For the best hyperparameter setting, \cref{mainalgs} (right) was repeated 50 times with different data-orders and $w_0$. In all figures any shaded area corresponds to 2 standard deviations around the mean as computed from the 50 runs.

\section{More details on direct risk bound minimization}
\label{app:drbm}

We evaluated the performance of a learning algorithm baed on optimizing a PAC-Bayes bound based on a data-dependent prior.
Our proposed algorithm gets nearly state-of-the-art performance and produces a valid and tight PAC-Bayes bound on risk.

Let $Q = \Normal{w}{\Sigma_{\alpha}}$ and $P = \Normal{w_{\alpha}}{\Sigma_{\alpha}}$.
The algorithm starts with the coupling and \prefix{} runs as before.
Then the \full{} run is replaced with SGD minimizing the PAC-Bayes bounds $\diffLPBB{S\setminus S_{\alpha}}{Q(\theta_Q)}{P}{\delta}$ with respect to the posterior mean $w$. Here $\diffLPBB{S\setminus S_{\alpha}}{Q(\theta_Q)}{P}{\delta}$ is the same bound as $\klLPBB{S\setminus S_{\alpha}}{Q(\theta_Q)}{P}{\delta}$ but with risk evaluate on a differentiable surrogate loss.
The procedure is outlined in \cref{mainalgs} (left).

Similarly as before, for each $\alpha$ we choose the learning rate and prior variance that yield the tightest bounds. For a fixed set of hyperparameters, we repeat the optimization 50 times.

The results on 4 different networks and 3 different datasets appear in
\cref{fig:directboundopt}.
The risk bounds and test errors drop dramatically with $\alpha$ up to $\alpha \approx 0.9$ for all the networks tested.
For MNIST and Fashion-MNIST, the momentum and batch size is the same as above. For CIFAR-10, the momentum is 0.9 and the batch size is 128. For MNIST and Fashion-MNIST, the \prefix{} run learning rate is 0.01; for CIFAR-10 it is 0.03. For all datasets, we sweep the direct bound optimization learning rate over $\{\SI{1e-6}, \SI{3e-6}, \SI{1e-5}, \dots, \allowbreak \SI{3e-3} \}$ and the prior variance over $\{\SI{1e-9}, \allowbreak \SI{3e-9}, \SI{1e-8}, \dots, \SI{3e-3} \} $.

\subsection{Comparison to PAC-Bayes by Backprop}

When $\alpha=0$, the setting of our direct bound optimization experiments aligns closely to that considered by \citet{rivasplata2019pac}: evaluating a PAC-Bayes bound-based learning algorithm using a prior centered at random initialization. This work reports a test error of $0.014$ and a risk bound of $0.023$ on MNIST with a 784--600--600--10 fully-connected network architecture, a Gaussian prior, and a PAC-Bayes bound expression similar to ours. Despite correspondence with the authors, we were unable to reproduce these results. For direct comparison, our $\alpha=0$ baseline results with the same network architecture are a mean test error of $0.116$ and a mean risk bound of $0.303$ over 10 random seeds. Using a data-dependent prior learnt with proportion $\alpha=0.7$ of the training data, this improves to a mean test error of $0.022$ and a mean risk bound of $0.031$ over 10 random seeds.

\section{Optimal prior variance}
\label{sec:oraclevar}

Our data-dependent priors do not attempt to minimize the variance component of the KL bound.
For a fixed $\Sigma_P$, the variance component in \cref{klgaussians} (see \cref{app:approx}) increases if posterior variance $\Sigma$ deviates from $\Sigma_P$.
When the prior is isotropic, our empirical study shows that the optimized posterior variance is also close to isotropic.
However, an isotropic structure may not describe the local minima found by SGD well.
We are thus also interested in a hypothetical experiment, where we allow the prior variance to be optimal for any given diagonal Gaussian $Q$.
While this produces an invalid bound, it reveals the contribution to the risk bound due to the prior variance.
Optimizing \cref{klgaussians} w.r.t. diagonal $\Sigma_P$ yields a prior $P^{\Sigma}_{\alpha}$ with optimal variance,
and $\KL{Q}{P^{\Sigma}_{\alpha}}$ expression reduces to
\begin{equation}\label{eq:oraclevariance}
\smash{
\textstyle 
\frac 1 2
\sum_{i=1}^p \log ( 1 + {( w^i-w^i_{\alpha})}/ {\sigma^2_i} ),
}
\end{equation}
where $\sigma^2_i$ is the $i^{\,\text{th}}$ component of the diagonal of $\Sigma$.

Computing these hypothetical bounds with $P^{\Sigma}_{\alpha}$ as a prior requires some minor modifications to  \cref{mainalgs} (right).
As in  \cref{mainalgs} (right), the posterior is set to $Q = \Normal{w_{S}}{\Sigma}$,
with a diagonal covariance matrix $\Sigma$ that is initialized to $\sigma^2_P I_{p}$. The prior $P$ is centered at $w^0_{\alpha}$, and the variance is automatically determined by the posterior variance.
The KL then takes the form stated in \cref{eq:oraclevariance}.
The \prefix{} run in \cref{mainalgs} (right) is followed by another SGD run minimizing $\klLPBB{S\setminus S_{\alpha}}{Q(\theta_Q)}{P}{\delta}$ with respect to diagonal covariance $\Sigma$.

\begin{figure}[t]
\centering
\includegraphics[width=.49\linewidth]{sgd_mnist_lenet_final}
\includegraphics[width=.49\linewidth]{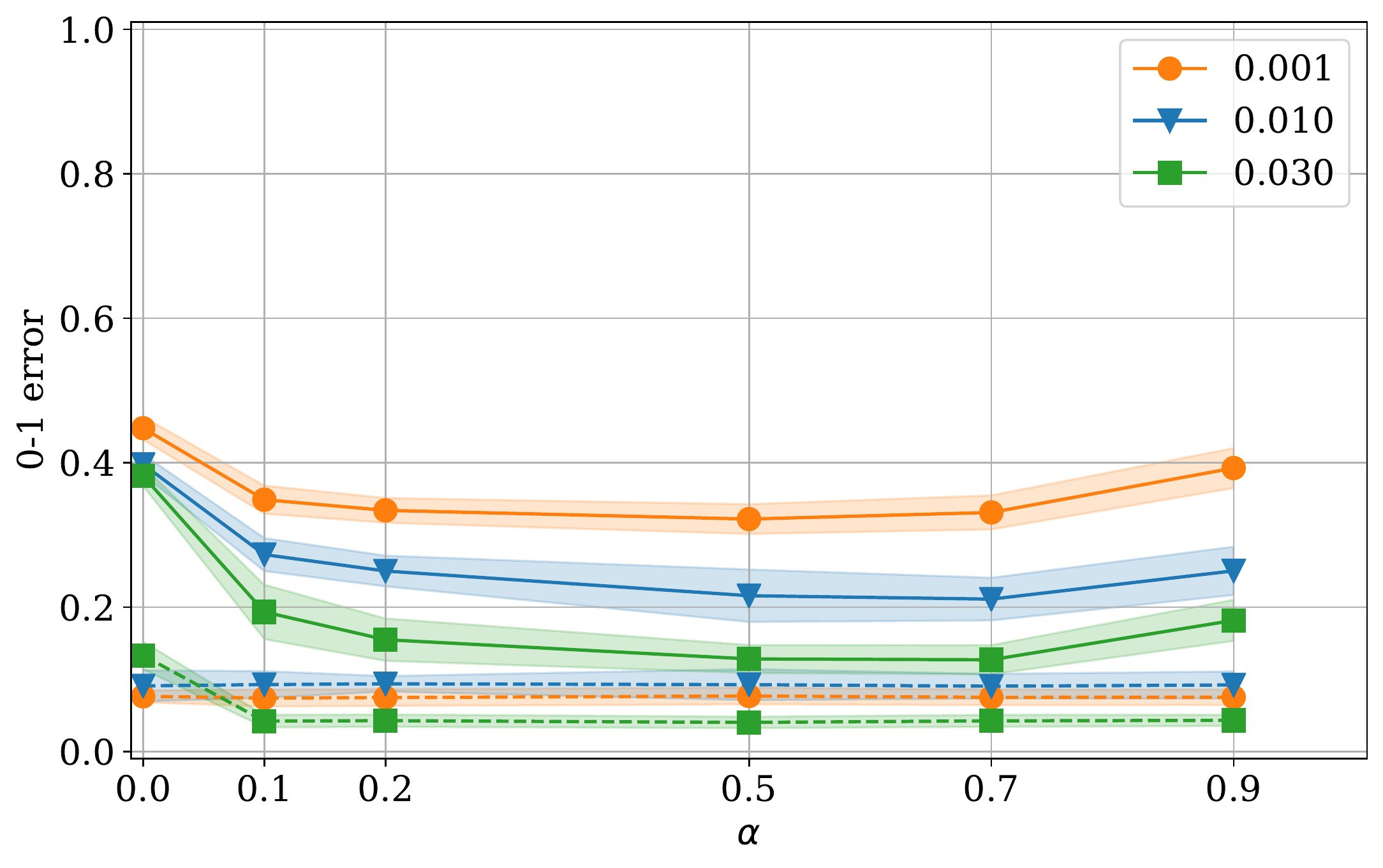}

\includegraphics[width=.49\linewidth]{sgd_fashion_mnist_lenet_final}
\includegraphics[width=.49\linewidth]{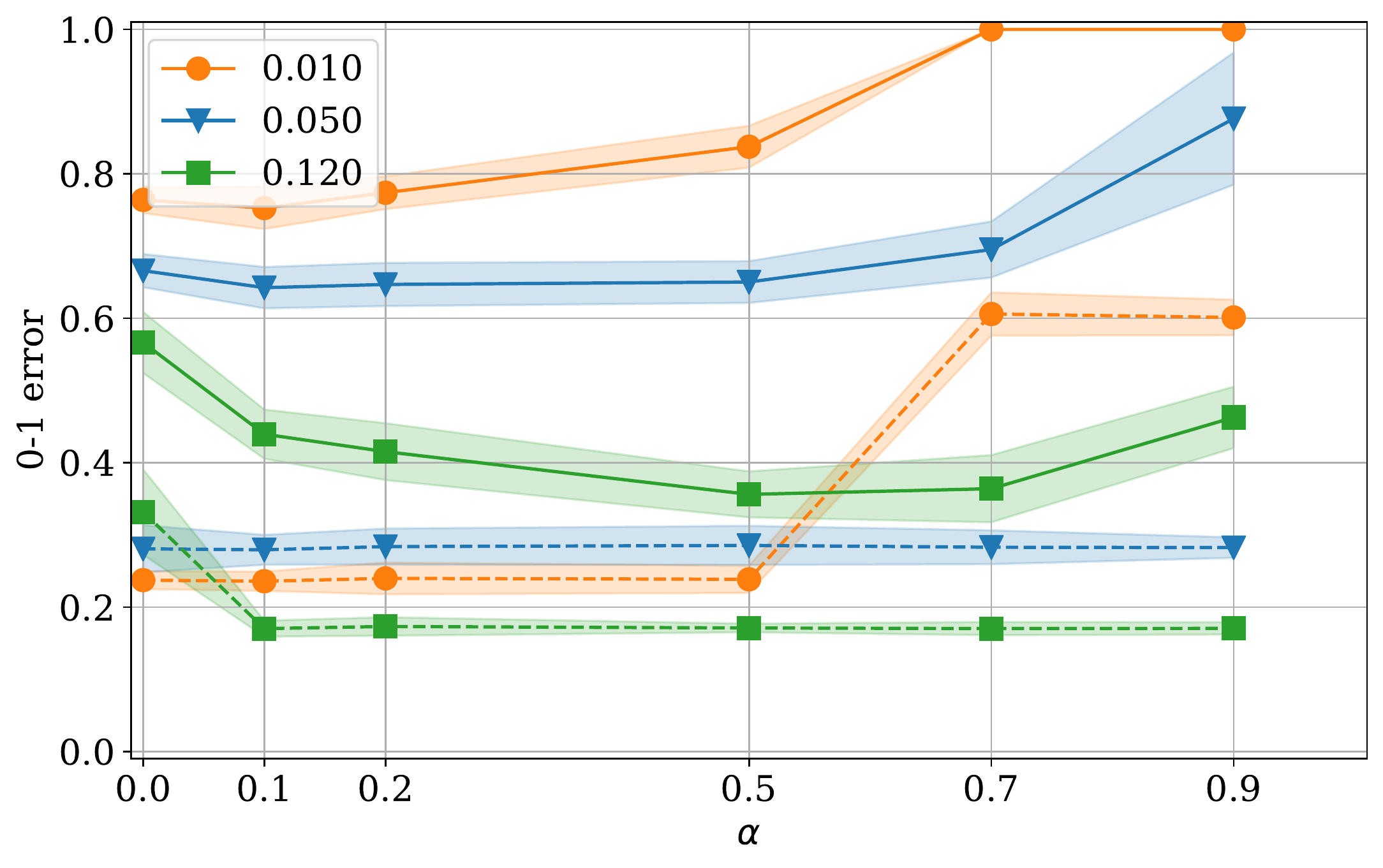}%

\includegraphics[width=.49\linewidth]{sgd_mnist_mlp_final}
\includegraphics[width=.49\linewidth]{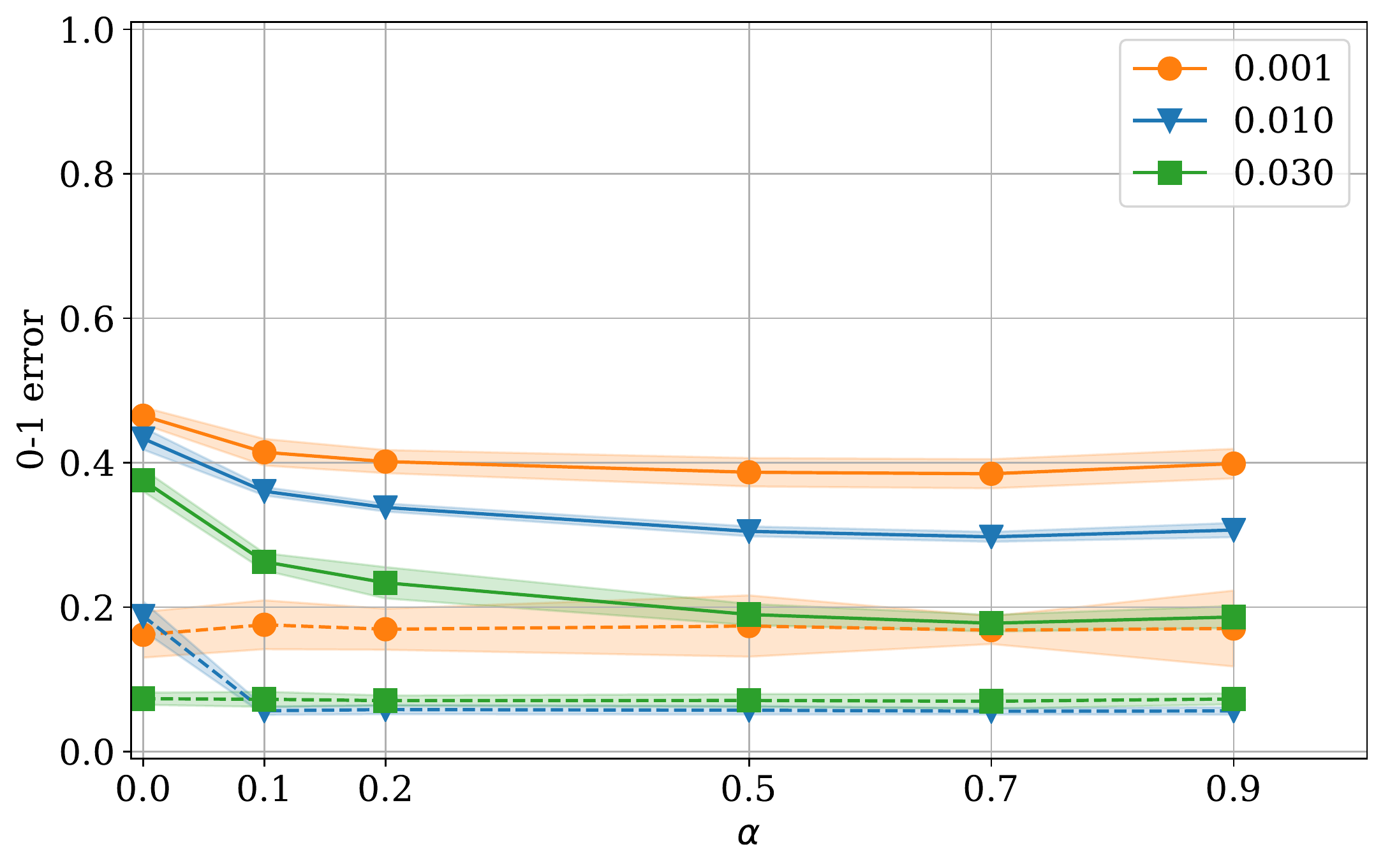}%

\caption{\textbf{Top row}: MNIST, LeNet-5; \textbf{middle row}: Fashion-MNIST, LeNet-5; \textbf{bottom row}: MNIST, FC;
\textbf{y-axis}: error-rate;
\textbf{x-axis}: fraction $\alpha$ of the data used by the \prefix{} run of SGD to predict the weights produced by the \full{} run of SGD;
\textbf{dashed lines}: test error;
\textbf{solid lines}: bound on the error of a Gaussian Gibbs classifier whose mean is the weights learned by the \full{} run of SGD and whose covariance has been optimized to minimize a PAC-Bayes risk bound;
\textbf{legend}: training error that was used as the stopping criterion for the \full{} run;
\textbf{left column:} test error and PAC-Bayes generalization bounds with isotropic prior covariance;
\textbf{right column:} hypothetical bounds with diagonal prior variance set to optimal. The improvement is seen only for low $\alpha$ values. At higher $\alpha$ values, the bounds are similar to the ones obtained with isotropic prior variance. The best test error bound on MNIST (approximately 11\%) is significantly better than the 46\% bound by \citet{Zhou18}.
}
\label{fig:boundandoracle}
\end{figure}

The results with the optimal prior covariance can be found in \cref{fig:boundandoracle}.
At $\alpha=0$, the optimal prior variance decreases the bound substantially. However, at larger values of $\alpha$, the effect diminishes. In particular, at the values of $\alpha$ that produce the lowest risk bound with a fixed isotropic prior variance, optimal prior variance makes little to no improvement. Interestingly, the optimized posterior variance remains close to isotropic. 

\begin{figure}[t]
\centering
\includegraphics[width=.49\linewidth]{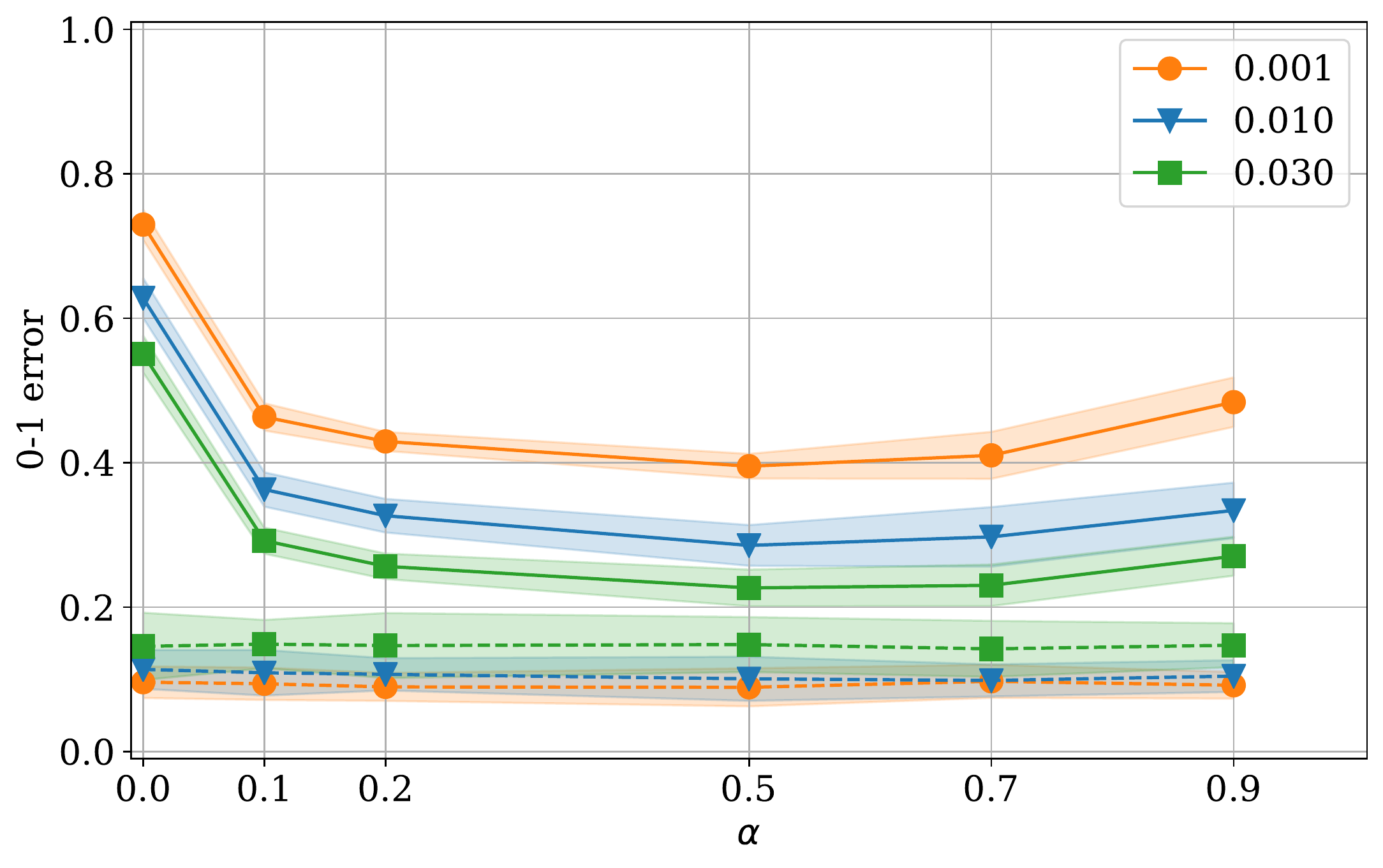}
\includegraphics[width=.49\linewidth]{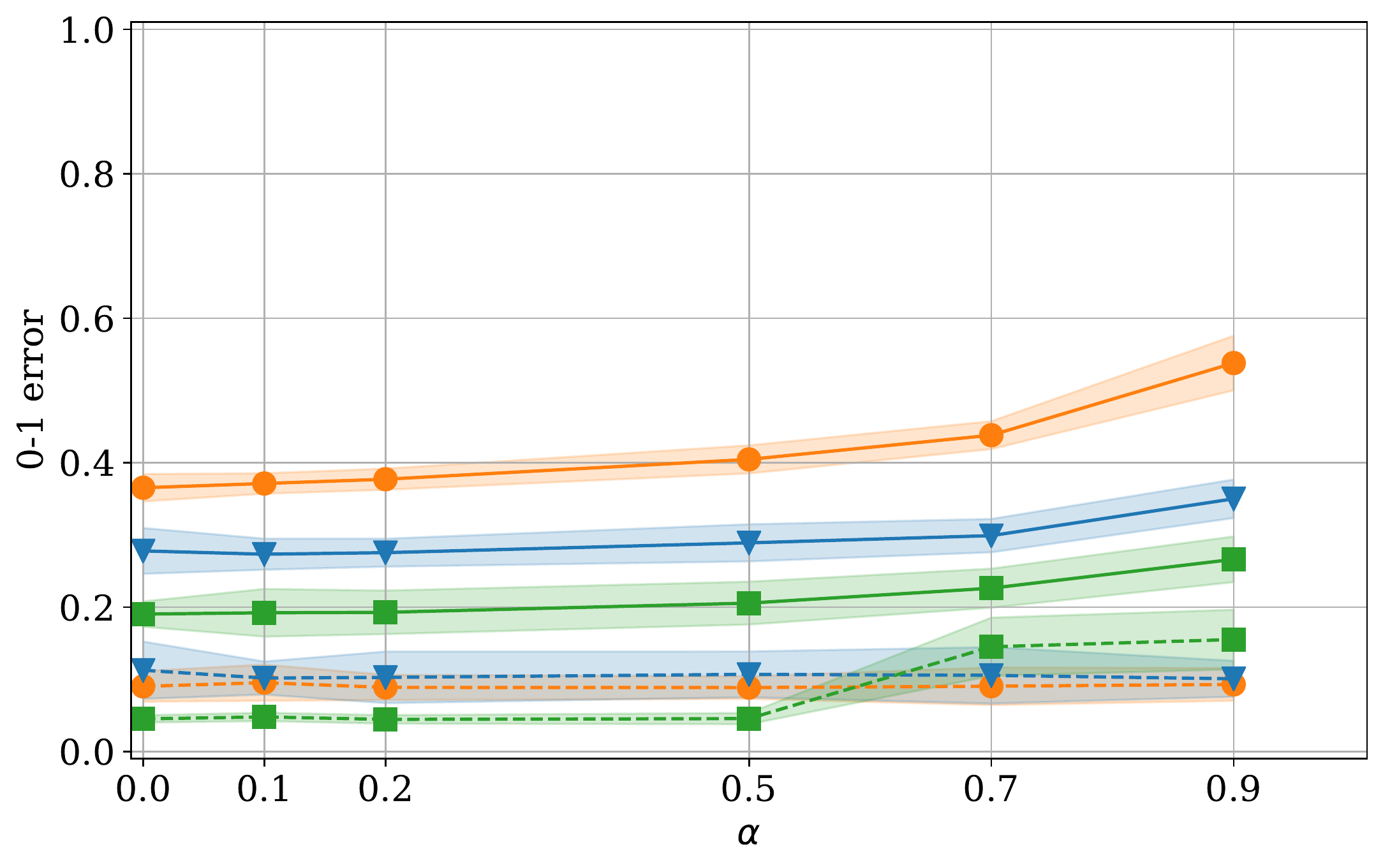}

\includegraphics[width=.49\linewidth]{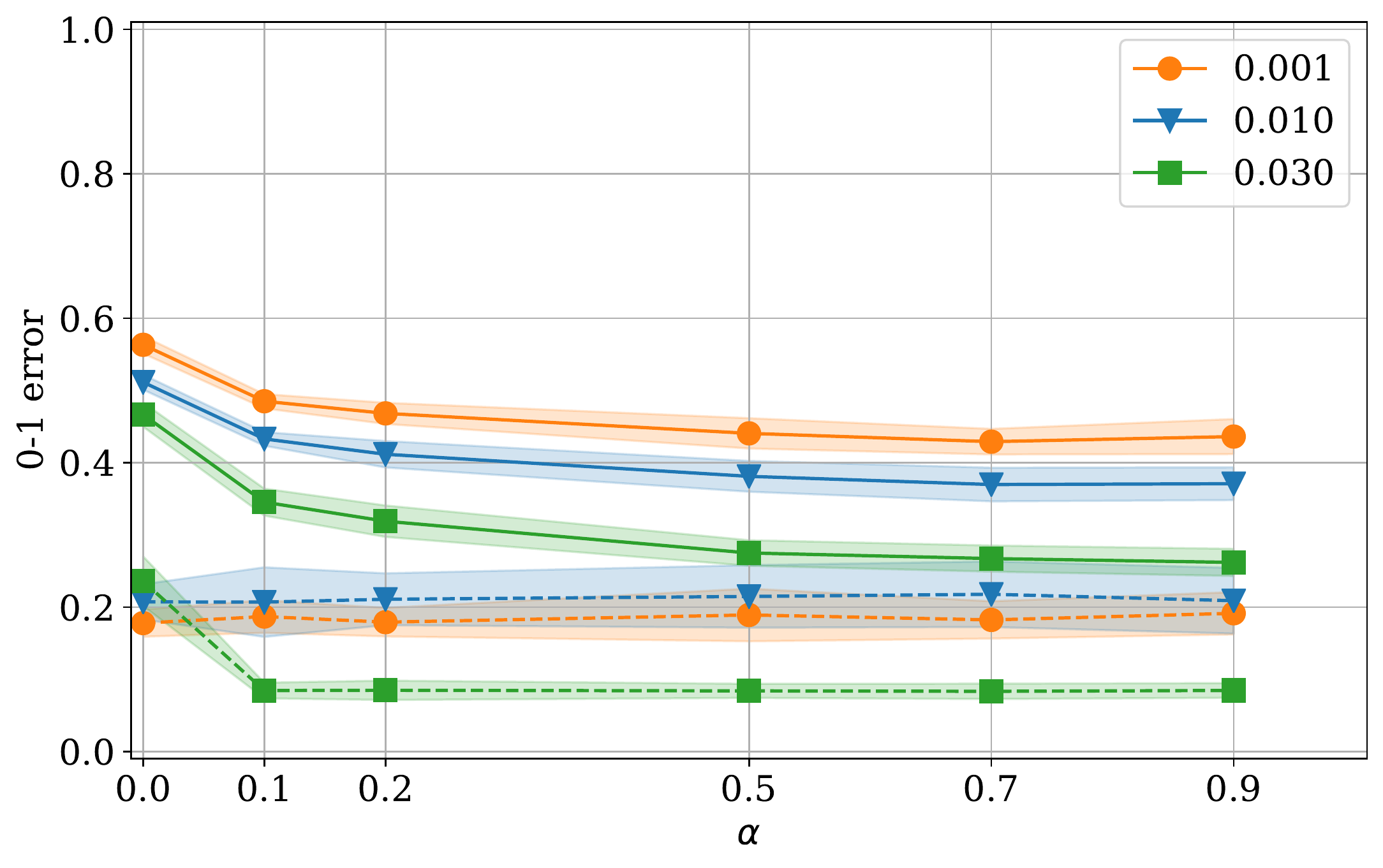}
\includegraphics[width=.49\linewidth]{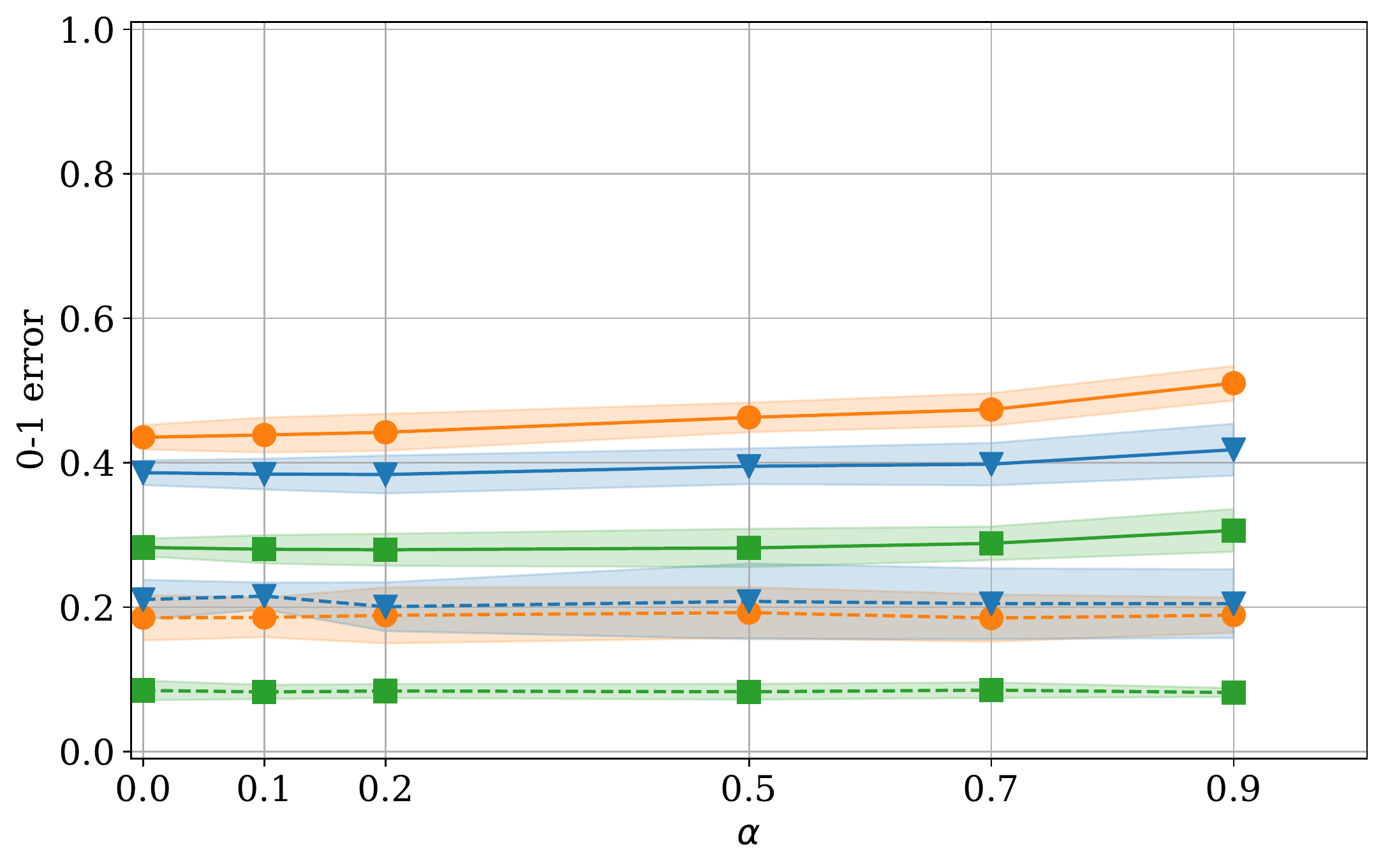}

\caption{
\textbf{Top row}: MNIST, LeNet-5; \textbf{bottom row}: MNIST, FC; \textbf{y-axis}: error-rate; \textbf{x-axis}: fraction $\alpha$ of the data used by the \prefix{} run of SGD to predict the weights produced by the \full{} run of SGD;
\textbf{left column}:
test error and PAC-Bayes error bounds with isotropic prior covariance using half of MNIST data;
\textbf{right column}: data and oracle prior bounds, where the prior is an isotropic Gaussian.
The oracle prior is approximated by using ghost samples. When using ghost samples, some improvement on the bounds is seen for small values of $\alpha$ (below 0.2). For large values of $\alpha$ (at around 0.9) and some stopping times, the bound with a data and oracle dependent prior is worse than with a data-dependent prior.
}
\label{fig:ghostbound}
\end{figure}

\section{Frequently Asked Questions}
\label{app:misconceptions}

\newcommand{\highl}[1]{\textbf{#1}}

\paragraph{Data-dependent priors are already a known heuristic for obtaining tight bounds, right?}
\citet{Amb07} proposed to use data-dependent priors and observed that they led to much tighter bounds than those produced by generic priors. 
However, these bounds were motivated by trying to approximate distribution-dependent oracle priors (producing so-called localized PAC-Bayes bounds \citep{langford2003microchoice,Catoni}). Indeed, in their work, the data-dependent prior is based on an estimate of the data-independent oracle prior. The bound contains a penalty for the error in this estimate, and so it will not be tighter in expectation. 

In contrast, in this work, we observe that data-dependent priors can be superior to distribution-dependent oracle priors. This is not folklore.
The effect we are observing is due to the fact that not all data are created equal.
As demonstrated by \cref{mainthm} or \cref{mainexample}/\cref{keyclaim}, using data that have a particularly strong a priori dependence on the posterior can dramatically tighten the risk bound. In our example, the initial data have greater dependence because of the decreasing step size. In the end, data-dependence is the difference between a vacuous  bound  (based on the distribution-dependent oracle prior) and a nonvacuous bound (based on a data-dependent oracle prior).

\paragraph{How does using a data-dependent prior compare to using held-out data in a, e.g., Chernoff bound?}
There is a critical difference: in the former, the posterior $Q(S)$ depends on the entire sample, $S$. In the latter, the held-out data would not be used by the posterior. The held-out bound \emph{cannot} be used to explain why generalization is occurring. It simply reports that generalization has occurred. In contrast, bounds based on data-dependent priors lend themselves to arguments in terms of distribution-dependent stability. We don't pursue the interpretation of these bounds here, though we discuss related issues in \cref{sec:sgdddpriors}.

\paragraph{What's the relationship between direct risk bound optimization (last subsection of \cref{sec:direct}) and the goal of explaining SGD?} This final subsection is not directed towards understanding SGD. It presents a novel learning algorithm, though the idea of minimizing a PAC-Bayes bound is a standard one. The use of a data-dependent prior and in particular this one based on a run of SGD on an initial segment of data is new.  

\paragraph{The paper studies minimizing high-probability PAC-Bayes bounds in expectation. Shouldn't we be using bounds on the expected generalization error?}
Both approaches are sensible. 
Note that a PAC-Bayes bound controls the generalization error in terms of the KL divergence between the posterior and the prior. (Other types of PAC-Bayes bounds exist, but we will focus on this standard setting here.) The posterior is data dependent and so the KL divergence is a random variable, in general. 
If we want to develop a tight bound, we want to minimize the KL divergence term, but since it is a random variable, there's no unique way to minimize it. 
In this work, we minimize the contribution of the (random) KL divergence by minimizing its expectation. Since the PAC-Bayes bound holds with high probability, it would also be interesting to minimize a tail bound on the KL divergence. We do not pursue that here, but it is interesting future work.

\paragraph{What is the meaning of (i) $\LPBB{\beta}{S \setminus S_J}{Q(S)}{P^*_{\PF}(S_J)}{\delta}$ and (ii) $\LPBB{\beta}{S}{Q(S)}{P^*_{\PF}}{\delta}$ 
 in \cref{mainthm}?} From \cref{pacbayeslinear}, we see $\LPBB{\beta}{S}{Q}{P}{\delta}$ is the linear PAC-Bayes bound on the risk of a posterior $Q(S)$, based on the estimate $\EmpRisk{S}{Q(S)}$, using the prior $P$. The terms above are, therefore,
 \begin{enumerate} 
 \item[(i)] 
 the linear PAC-Bayes bound on the risk of $Q(S)$, based on the estimate $\EmpRisk{S \setminus S_J}{Q(S)}$, using the (data-dependent) prior $P^*_{\PF}(S_J)$; and
\item[(ii)] 
 the linear PAC-Bayes bound on the risk of $Q(S)$, based on the estimate $\EmpRisk{S}{Q(S)}$, using the (data-independent) prior $P^*_{\PF}$.
\end{enumerate}
Therefore, the theorem is telling us when a linear PAC-Bayes bound can be improved using a data dependent prior. The $\Psi$ term in \cref{keyclaim} can be interpreted in the same way.

\paragraph{There are a wide variety of PAC-Bayes bounds. Why do you use linear PAC-Bayes bounds and do your findings generalize to other types of PAC-Bayes bounds?}
Our focus on linear PAC-Bayes bounds allows us to simplify the analysis considerably. Indeed, by the linearity of expectation, the expected value of the linear PAC-Bayes bound depends on the expected value of the KL term, which then gives us the connection to mutual information. 

Regarding other styles of PAC-Bayes bounds, by Jensen's inequality, priors that minimize the expected value of the KL terms will lead to upper bounds on the classic sqrt-style PAC-Bayes bound. Minimizing an upper bound provides a weaker but still valid approach to controlling such bounds. There is also a connection between linear PAC-Bayes bounds and nonlinear PAC-Bayes bounds, such as that derived in \cref{app:inverting}. In particular, from the logic that leads to \cref{optbound}, we see that optimizing the value of $\beta$ leads to a bound of the same form as \cref{varklbound}. What we can glean from this correspondence is that the argument in \cref{keyclaim} is choosing the value of $\beta$ that is optimal in expectation.

Note that $\beta$ must be chosen independently of the data used in the estimate of the risk. However, one can consider a discrete range of beta values, derive a data-dependent prior (and then linear PAC-Bayes bound) for each such value, and then combine these bounds (i.e., taking the tightest one) using a union bound argument. The result is a nonlinear bound, though the final bound is no longer being minimized in expectation. To optimize a nonlinear bound directly, the easiest approach may be to control the tails of the KL term. We think this is an interesting avenue for future work.

\paragraph{Which is more important/fundamental: \cref{mainthm} or \cref{mainexample}/\cref{keyclaim}?}
We think these results enhances each other, and that they are both important.

When we showed colleagues \cref{keyclaim} alone, they were then eager to see a general characterization of when data-dependence led to tighter bounds. \Cref{mainthm} provides necessary and sufficient conditions. As is the case with necessary and sufficient conditions, the result simply presents a different, but equivalent, perspective. In this case, it shows how the superiority of a data-dependent prior comes from the relative values of the ``information rate gain'' and the ``excess bias''.
While \cref{mainthm} is immediate from definitions, we believe it provides guidance as to how to choose $J$.
In particular,  \cref{mainexample} and \cref{keyclaim} demonstrate that one may arrive at much tighter bounds 
by identifying samples $J$ that have a priori high dependence on the posterior. 

\paragraph{How should one choose the subset, $J$, of data used to build the data-dependent prior?}

We believe that \cref{mainthm} provides guidance: we need the information rate gain
 to exceed (some multiple of) the excess bias plus a variance term. 
 
The variance term will be quite small unless the number of data used in the prior, $m$, is quite large. 
Focusing then on the information rate gain, we will maximize this term
if we build our prior using samples that exhibit strong a priori dependence with the final posterior.
This will produce a tighter bound, provided that the samples that we leave to use in the risk estimate  are not too biased.

In \cref{mainexample}, 
there is strong dependence with the initial data because the step size is largest at the start.
At the same time, the learned weights still produce good predictions for all the data. 
The excess bias term captures the effect of removing this data from the risk estimate, which is sufficiently small in this case.

\paragraph{Is your goal to improve the generalization of SGD?} Our goal 
is \emph{not} to improve the generalization of the SGD algorithm, but to improve our ability to simultaneously (i) choose posteriors $Q$ that closely approximate the performance of SGD and (ii) derive tight generalization bounds for these posteriors. This hope is to shed light on SGD itself, eventually. As the paper argues, the roadblock here is (ii)---in particular, the priors in the KL terms in PAC-Bayes bounds are too far from the posteriors to yield numerically tight bounds. The final two paragraphs of the paper describe a learning algorithm, but this is simply a short aside, and not the main focus of the paper.

\paragraph{When using ghost samples for the prior, shouldn't one compare to the same algorithm that also has access to the ghost samples? }
No. We would agree with this sentiment if we were designing a learning algorithm or model selection method. However, we are instead probing the generalization properties of posteriors concentrated around the weights learned by SGD. Our experiment reveals that the advantage of distribution-dependence (provided via ghost samples) may be dwarfed by the advantage of data-dependence. Note that the posterior $Q(S)$ does not depend on either the ghost samples or prior.

Section 3 rigorously identifies a setting where data-dependence is required for nonvacuous (linear PAC-Bayes) bounds. In Section 4, Fig. 3, we turn to the question of teasing apart how much data- and distribution- dependence helps. Ghost samples provide distribution-dependence. Section 4.1 shows how to use ghost samples to estimate the optimal Gaussian prior mean (for a fixed variance). The curve with ghost data is therefore an estimate of the \emph{actual} tradeoff in $\alpha$. The curve without ghost data is indeed heuristic. The fact that this heuristic no-ghost-data curve nearly matches highlights a convenient empirical fact: we can maybe ignore ghost data for this class of Gaussian priors.

 \ \\
\end{document}